\documentclass[runningheads]{llncs}
\usepackage{eccv}

\usepackage{eccvabbrv}

\usepackage{graphicx}
\usepackage{booktabs}
\usepackage{adjustbox}
\usepackage[table]{xcolor}
\usepackage{multirow}
\usepackage{subcaption}  
\usepackage{makecell}

\usepackage[accsupp]{axessibility}  

\usepackage{hyperref}

\usepackage{orcidlink}

\begin{document}

\title{LoRC: Detecting AI-Generated Images via Low-Rank Collapse in Semantic Residuals} 

\titlerunning{LoRC}

\author{
Haozhen Yan\inst{1}\textsuperscript{\#}\orcidlink{0009-0001-2223-8402} \and
Ruoxin Chen\inst{2}\textsuperscript{\#}\orcidlink{0000-0001-8729-3034} \and
Jiahui Zhan\inst{1}\orcidlink{0009-0005-6721-6740} \and
Bo Wang\inst{2}\orcidlink{0009-0004-9426-8741} \and  \\
Youchang Xiao\inst{2}\orcidlink{0009-0008-9975-7703} \and
Shouhong Ding\inst{2}\orcidlink{0000-0002-3175-3553} \and 
Liqing Zhang\inst{1}\orcidlink{0000-0001-7597-8503} \and \\
Taiping Yao\inst{2}\textsuperscript{*}\orcidlink{0000-0002-2359-1523} \and 
Jianfu Zhang\inst{1}\textsuperscript{*}\orcidlink{0000-0002-2673-5860}
}

\authorrunning{Yan et al.}

\institute{Shanghai Jiao Tong University, China \and
Tencent Youtu Lab, China\\ 
\email{\{orion810, c.sis\}@sjtu.edu.cn, taipingyao@tencent.com}}

\maketitle

\begingroup
\renewcommand{\thefootnote}{\#}
\footnotetext[1]{These authors contributed equally.}
\endgroup

\begingroup
\renewcommand{\thefootnote}{*}
\footnotetext[1]{Corresponding authors.}
\endgroup

\begin{abstract}
  Modern generators faithfully model macroscopic semantics, producing synthetic images that appear highly realistic. Consequently, decisive forensic cues reside in subtle non-semantic visual discrepancies. 
  To reveal these cues, we revisit AIGI detection from a geometric perspective and identify an architecture-agnostic signature. Specifically, modern generators exhibit low-rank collapse (\textit{i.e.}, rank degeneracy) in the semantic-residual orthogonal subspace while largely preserving the dominant semantic direction. 
  This structural flattening consistently emerges during the final decoding stage, forming a shared bottleneck across diverse generator architectures.
  Motivated by this signature, we propose \textbf{LoRC}, a framework that decouples semantic dominance to capture the collapsed residual geometry induced by the generative decoding bottleneck.  
  Our method improves accuracy by an average of 7.0\% across multiple benchmarks and achieves 97.0\% accuracy on 39 unseen generators. 
  These results demonstrate strong cross-model generalization and robustness, making LoRC a reliable approach for AIGI detection in complex real-world environments.
  \keywords{AI-Generated Image Detection \and Low-Rank Collapse \and Orthogonal Projection}
\end{abstract}

\begin{figure}[tb]
  \centering
  \includegraphics[width=\linewidth]{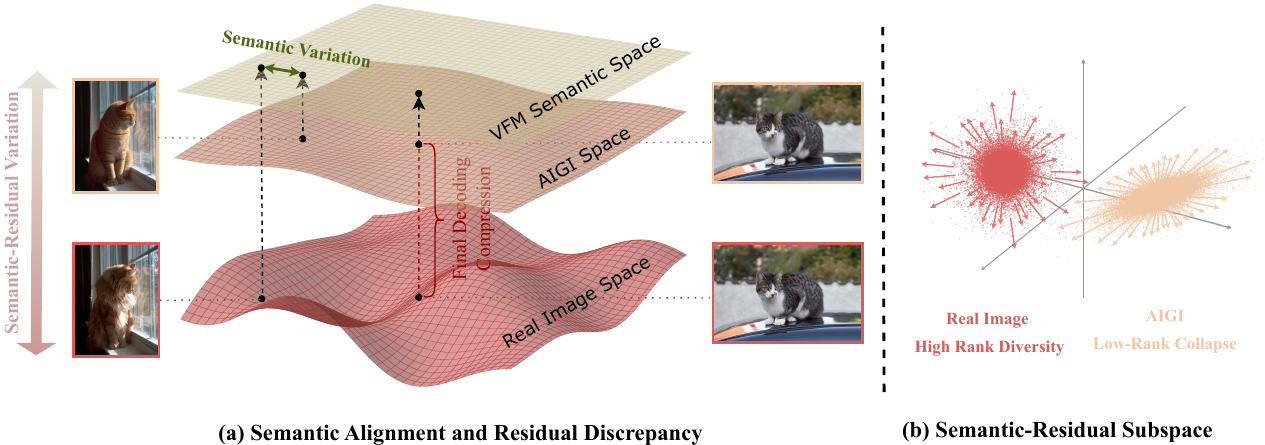}
  \caption{
    \textbf{Geometric perspective of AI-Generated Image (AIGI) detection.} We illustrate the structural relationship between real images, AIGIs, and the semantic space of vision foundation models (VFMs). While real images reside on an intricate, high-dimensional manifold (bottom), modern generative models are increasingly aligning with them in the dominant semantic space (top), causing the \textit{Semantic Variation} to shrink significantly. However, in the orthogonal semantic-residual dimension (vertical direction), the discrepancy between AIGIs and real images becomes correspondingly more pronounced. Our empirical analysis reveals that the final decoding stage of generative models acts as a severe information bottleneck, inducing a systematic \textbf{low-rank collapse} that forces the complex real manifold to flatten into a constrained AIGI space (middle). The resulting low-rank effect can serve as a universal forensic signature. 
  }
  \label{fig:teaser}
\end{figure}

\section{Introduction}
\label{sec:intro}
The widespread availability of AI-generated images (AIGIs)~\cite{goodfellow2014generative,ho2020denoising} raises growing concerns about authenticity and trust in digital content, driving the need for reliable detectors that distinguish AIGIs from real images. 
The rapid evolution of generative models further complicates detection by exacerbating cross-model generalization challenges, particularly in zero-shot settings involving unseen generators.
From a manifold learning perspective, natural images reside on a high-dimensional and complex real manifold that modern generators attempt to approximate.
With recent advances, generated images have become increasingly \emph{semantically} faithful, often matching real images in object identity and global layout (Fig.~\ref{fig:teaser}).
This is consistent with the coarse-to-fine nature of many generation pipelines~\cite{tian2024visual,KarrasALL18}, where global structure is stabilized early and later decoding mainly refines fine details.
Empirical evidence further supports this observation. As shown in Fig.~\ref{fig:genimage}, generated images from diverse architectures consistently exhibit stronger semantic alignment but weaker semantic-residual energy compared with real images.
As a result, the semantic dimension offers diminishing discriminative power for AIGI detection, especially under cross-model shifts and in zero-shot settings.
Consequently, the remaining evidence is pushed into the semantic-residual dimensions, where subtle non-semantic imperfections persist (\textit{e.g.}, texture naturalness and local physical consistency).

This observation naturally raises a critical question: \textbf{does a universally exploitable and systematic discrepancy actually exist within these semantic-residual dimensions?} The answer is affirmative.
To reveal this discrepancy, we analyze deep visual representations by geometrically decoupling them into a primary semantic direction and its orthogonal residual subspace.
As illustrated in Fig.~\ref{fig:vae_recons2}, controlled reconstructions of real images induce highly directional feature shifts that consistently reduce residual magnitude while leaving the semantic direction largely unchanged.
Furthermore, the variance spectrum of these shifts (Fig.~\ref{fig:pca}) reveals a striking phenomenon within the residual space: a systematic \textbf{Low-Rank Collapse}, where the discrepancy concentrates on only a few dominant dimensions.
Specifically, the reconstruction of real images induces feature shifts that are severely compressed and strictly directional in the semantic-residual dimensions, while introducing no significant variations along the semantic direction.
Crucially, our empirical analysis in Fig.~\ref{fig:pca_matrix} reveals two key properties of the low-rank collapse, highlighting it as a generalizable and robust generative signature.
\textbf{First, it is cross-architectural.}
This phenomenon appears at the \emph{final decoding stage} across fundamentally different paradigms, including diffusion and autoregressive models, indicating that the final generative mapping acts as a shared information bottleneck.
\textbf{Second, it is inherently robust.}
Because the deviation arises from the geometric structure of the feature manifold, it is largely insensitive to superficial perturbations such as JPEG compression.
Together, these properties suggest that low-rank collapse transcends model-specific artifacts and provides a reliable signal for real-world forgery detection.

Motivated by these geometric insights, we propose LoRC (Leveraging Low-Rank Collapse), a framework that shifts AIGI detection from fragile spatial artifacts to robust manifold geometry. Specifically, LoRC isolates and captures this generative signature through a three-stage pipeline:
First, to neutralize dominant semantic alignment, we introduce \textbf{Semantic Decomposition}. Using a frozen vision foundation model (VFM), we treat the global \texttt{[CLS]} token as a semantic anchor and project patch tokens onto its orthogonal complement. This isolates the semantic-residual subspace.
Second, because VFM embeddings are high-dimensional, the subtle low-rank collapse may be obscured by irrelevant feature variations. To address this, we introduce a \textbf{Low-Rank Attention Block}. This module restricts residual features to a rank-$r$ bottleneck before attention computation, encouraging the network to focus on the collapsed semantic-residual dimensions while suppressing noise.
Finally, to explicitly shape the manifold geometry, we introduce an auxiliary \textbf{Subspace Separation Loss (SSL)}. By minimizing the inner product between normalized covariance descriptors of real and fake residuals, this loss enlarges the geometric distance between the high-rank natural manifold and the collapsed synthetic subspace.
By modeling the architectural limits of modern generators rather than superficial pixel cues, our framework achieves strong robustness and transferability. Extensive evaluations validate this geometric advantage: LoRC improves performance by an average of 7.0\% across multiple benchmark datasets. When evaluated on recent unseen generators spanning different architectural paradigms, our method achieves an accuracy of 97.0\%. These results demonstrate that LoRC provides a robust and reliable solution for AIGI detection in unconstrained real-world scenarios.

\section{Related Works}
\subsection{Early Low-level Detection}
Early AIGI detectors primarily exploit low-level artifacts introduced by the generation pipeline.
Zhang~\cite{zhang2019detecting} shows that GAN images can leave recognizable frequency-domain traces (e.g., spectral duplication) that serve as detection cues. However, later studies~\cite{Corvi_2023,karageorgiou2024any} report that such frequency patterns are substantially weaker for diffusion-generated images.
Building on this line, subsequent works~\cite{tan2024frequency, chu2024fire, karageorgiou2024any} leverage frequency-domain artifacts as auxiliary signals to strengthen discrimination.
NPR~\cite{tan2024rethinking} further observes that generators can introduce upsampling artifacts.
However, these low-level cues often exploit spurious, non-causal factors (\textit{e.g.}, image format) and are fragile under common degradations, which limits cross-model generalization and reduces detection accuracy.

\subsection{Dataset Bias and Alignment}
Recent studies~\cite{grommelt2024fake} show that detectors can inadvertently exploit non-causal dataset biases (e.g., file format, resolution, compression settings, or acquisition pipelines) as spurious cues, leading to poor cross-domain generalization.
A common remedy is to align real and synthetic data in both content and statistical distributions. Diffusion-reconstruction approaches map an input image to a reconstruction generated by a pretrained diffusion model and treat the reconstruction residual as forensic evidence~\cite{wang2023dire}, or compare real–real and fake–fake reconstruction pairs to improve robustness~\cite{chen2024drct}.
Other works incorporate semantic-guided regeneration~\cite{yu2024semgir} or frequency-guided reconstruction errors~\cite{chu2024fire} to improve alignment quality and reduce reconstruction bias. Distribution alignment has also been explored through inpainting and VAE-based reconstruction, as demonstrated in B-Free~\cite{guillaro2024bias}, AlignedForensics~\cite{rajan2025aligned}, REM~\cite{liu2025beyond} and M2EA~\cite{liu2026m}.
Beyond pixel-space alignment, DDA~\cite{chen2025dual} observes that alignment may introduce additional mismatch in the frequency domain, and therefore proposes dual alignment in both pixel and frequency spaces to suppress high-frequency bias. 
Overall, these works highlight the importance of bias mitigation and distribution alignment. However, they primarily operate at the data level and leave unanswered whether a more structural, cross-architectural forensic signature exists.

\subsection{Semantic Decoupling}
With the rise of foundation vision and vision-language models, frozen generic representations have become strong priors for in-the-wild detection. UnivFD~\cite{ojha2023towards} demonstrates that frozen features from CLIP-like models paired with a simple classifier substantially improve generalization to unseen generators. 
Simplicity Prevails\cite{zhou2026simplicity} further argues that modern VFM representations may already encode useful forensic cues.
For AIGI detection, the key forensic cues can differ from the ``what is depicted'' semantic cues that VFMs are trained to capture. When real and generated images are highly aligned in semantics, dominant semantic components can mask or suppress subtle but critical forensic signals.
Consequently, recent works investigate semantic-forensic decoupling and efficient model adaptation.
NS-Net\cite{yan2025ns} builds a semantic null-space and projects visual features into it to eliminate semantic interference.
VIB-Net\cite{zhang2025towards} imposes an information bottleneck to compress and discard prior semantic knowledge, encouraging minimal sufficient representations.
From an architectural perspective, Liu et al. \cite{liu2026exploiting} introduce a taxonomy of image generators based on their final architectural components and observe that the final component can imprint transferable traces useful for cross-generator generalization.
Effort\cite{yan2025effort} performs SVD on the parameter space and fine-tunes only the smallest components to learn forensic cues efficiently.
While these studies attempt to isolate forensic cues within specific parameter subspaces or architectural traces, their geometric entanglement with semantic structures remains systematically unexplored, motivating explicit subspace decoupling and geometric analysis.

\section{Methodology}
\subsection{Motivation}
\noindent \textbf{Semantic Alignment and Residual Discrepancy.} 
Recent advances in photorealistic image synthesis have significantly improved instruction adherence and perceptual realism. Generated images now closely align with real photographs in high-level structure, object identity, and relationships. 
However, this evolution is asymmetric and non-synchronous: high-level semantics are easier to learn and strongly constrained by high-dimensional conditioning signals (\textit{e.g.}, text), whereas low-level image formation cues remain more entangled and harder to fit. 
Consequently, this uneven progress pushes the remaining real–synthetic gap into subtle \textbf{semantic-residual} dimensions beyond content and composition.
Although modern vision foundation models such as CLIP~\cite{radford2021learning} and DINO~\cite{caron2021emerging,simeoni2025dinov3} provide powerful priors, their embeddings are optimized for semantic recognition. They capture what an image depicts, rather than the subtle process-dependent cues that distinguish real images from synthetic ones. 
As the remaining discrepancy becomes confined to semantic-residual dimensions, naive reliance on global semantic embeddings may suppress these weak forensic signals and introduce semantic interference.

This suggests that decisive forensic evidence does not reside in global semantic embeddings, but in process-dependent residuals left by the generator. 
We therefore revisit the underlying generation mechanism. 
A recent study~\cite{liu2026exploiting} suggests that diverse generators share a final decoding or rendering stage that maps intermediate representations to the final image (\textit{e.g.}, a VAE decoder, a super-resolution diffusion head, or a VQ de-tokenizer). This stage may consistently imprint transferable synthetic traces across model families.
Building on this insight, our geometric analysis reveals that fundamentally different paradigms, including diffusion and autoregressive models, exhibit the same structural flaw. 
Their final decoding stage acts as a severe information bottleneck that induces systematic \textbf{low-rank collapse} within the semantic-residual subspace. 
This intrinsic geometric degradation provides a universal, architecture-agnostic signal for generalized AIGI detection.

\noindent \textbf{Geometric Decoupling in Feature Space.}
Given an input image $\mathcal{I} \in \mathbb{R}^{H \times W \times 3}$, we extract deep features from a frozen VFM $f_\phi$, obtaining a global token and patch tokens $\{\mathbf{c}, \mathbf{X}\} = f_\phi(\mathcal{I})$, where $\mathbf{c} \in \mathbb{R}^{D}$ denotes the \texttt{[CLS]} token and $\mathbf{X} \in \mathbb{R}^{N \times D}$ contains $N$ spatial patch embeddings. We normalize the frozen \texttt{[CLS]} token as $\hat{\mathbf{c}} = \mathbf{c} / \|\mathbf{c}\|_2$, which defines the principal semantic direction in feature space. We then perform an orthogonal decomposition of the patch feature map:
\begin{equation}
    \mathbf{X}_{sem} = \mathbf{X} (\hat{\mathbf{c}}\hat{\mathbf{c}}^\top), 
    \quad
    \mathbf{X}_{res} = \mathbf{X} (\mathbf{I} - \hat{\mathbf{c}}\hat{\mathbf{c}}^\top),
    \label{eq:res}
\end{equation}
where $\mathbf{I} \in \mathbb{R}^{D \times D}$ is the identity matrix. Here, $\mathbf{X}_{sem}$ captures the dominant global semantic alignment, while $\mathbf{X}_{res}$ isolates the semantic-residual subspace that defines the remaining discrepancy between real and synthetic images.

\begin{figure}[tb]
    \centering
    \begin{subfigure}{0.49 \linewidth}
        \includegraphics[width=\linewidth]{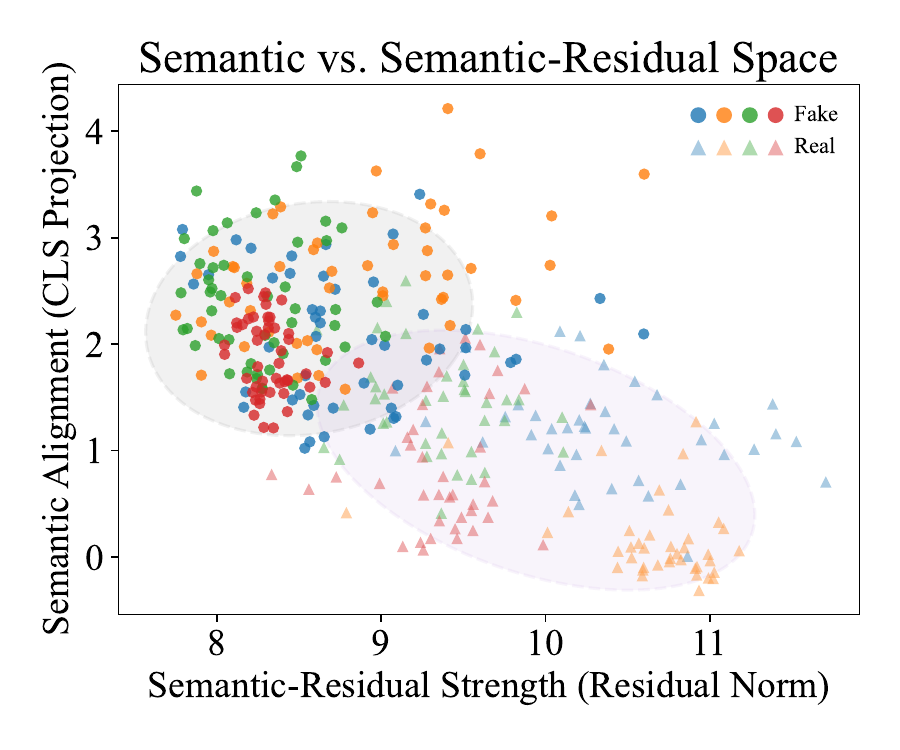}
        \caption{Feature distributions in the decoupled semantic-residual space on the GenImage. Colors denote categories. The consistent upper-left clustering of fake samples exposes a systemic generative bottleneck: semantic over-alignment coupled with attenuated residual energy.}
        \label{fig:genimage}
    \end{subfigure}
    \hfill
    \begin{subfigure}{0.49 \linewidth}
        \includegraphics[width=\linewidth]{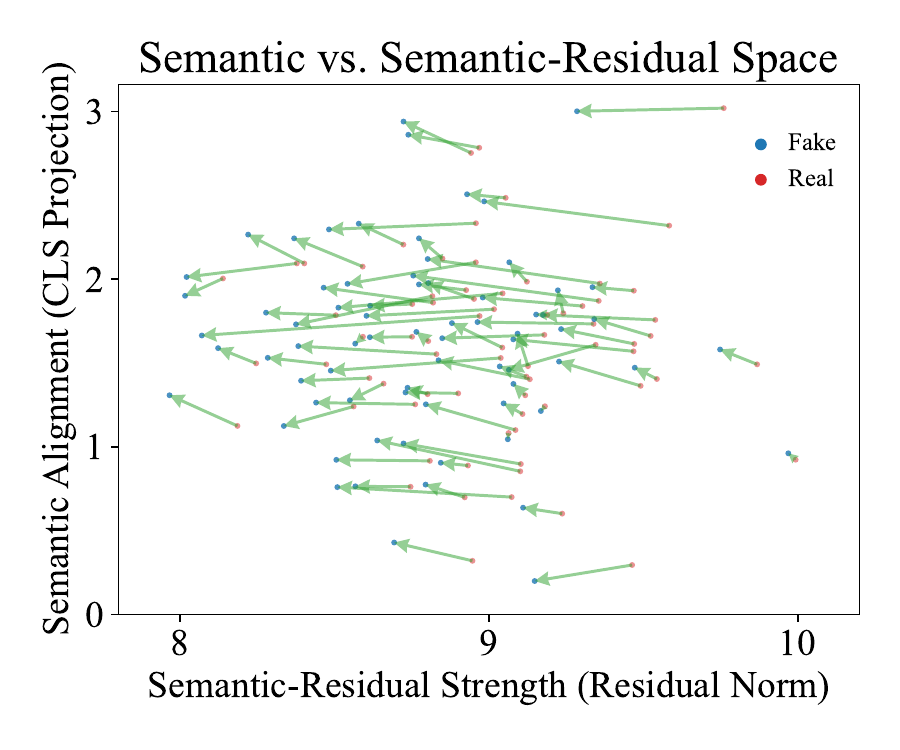}
        \caption{Feature trajectories of paired real and reconstructed samples. Directed edges show systematic leftward and slightly upward shifts, demonstrating that the terminal decoding stage induces residual attenuation and semantic over-alignment.}
        \label{fig:vae_recons2}
    \end{subfigure}
    \caption{Decoupled semantic–residual feature space on GenImage (a) and SD v1.5 VAE reconstructions (b), showing a consistent generative bottleneck: semantic over-alignment with attenuated residual energy.}
\end{figure}

\noindent \textbf{Systematic Semantic–Residual Shift.} 
Using this geometric formulation, we analyze feature distributions across diverse generative architectures using GenImage~\cite{zhu2023genimage}, a benchmark containing real images and their synthetic counterparts from eight generators. 
For each image $k$, we summarize the decomposed feature map using two scalar statistics that measure semantic alignment and residual magnitude:
\begin{equation}
Y^{(k)}=\frac{1}{N}\sum_{i=1}^{N}\mathbf{x}_i^{(k)\top}\hat{\mathbf{c}}^{(k)}, 
\qquad
X^{(k)}=\frac{1}{N}\sum_{i=1}^{N}\left\lVert \mathbf{x}_{res,i}^{(k)}\right\rVert_2 ,
\end{equation}
where $\mathbf{x}_i^{(k)}$ and $\mathbf{x}_{res,i}^{(k)}$ denote the $i$-th patch tokens of $\mathbf{X}^{(k)}$ and $\mathbf{X}_{res}^{(k)}$.
As shown in \cref{fig:genimage}, where colors indicate image categories, generated images (circles) from eight architectures consistently cluster in the upper-left region, exhibiting higher semantic alignment and smaller residual norms. In contrast, real images (triangles) spread toward regions with larger residual norms. This observation reveals two key findings: (i) modern generators tend to over-align with the dominant semantic direction, likely because macroscopic structures are easier to model, and (ii) semantic-residual energy is systematically attenuated in generated images.

\noindent \textbf{Semantic-Controlled Reconstruction.} 
While the previous analysis reveals a universal trend, the real and generated samples share only categorical labels. Moreover, modern image generation pipelines are complex, which may obscure the origin of the observed compression. To strictly align the semantic content and isolate the bottleneck effect, we sample MS-COCO\cite{coco} images and reconstruct them exclusively via the Stable Diffusion V1.5\cite{rombach2022high} VAE. Using these semantically matched pairs, we examine whether the terminal decoding stage alone can induce systematic attenuation of spatial residuals.
As shown in \cref{fig:vae_recons2}, we visualize the geometric transitions of these paired samples, where each directed edge represents the feature shift from a pristine real image to its VAE-reconstructed counterpart, denoted as the \textbf{feature delta ($\Delta$)}. 
The directed shifts reveal two consistent patterns: (1) The displacement vectors are predominantly oriented horizontally toward the left, indicating a strong attenuation of the semantic-residual norm. (2) Along the vertical semantic axis, most shifts show a slight upward tendency toward stronger semantic alignment. By controlling the visual content, these paired trajectories provide strong evidence that the final decoding component alone can introduce a systematic structural deviation in feature space.

\begin{figure}[tb]
  \centering
  \includegraphics[height=5.5cm]{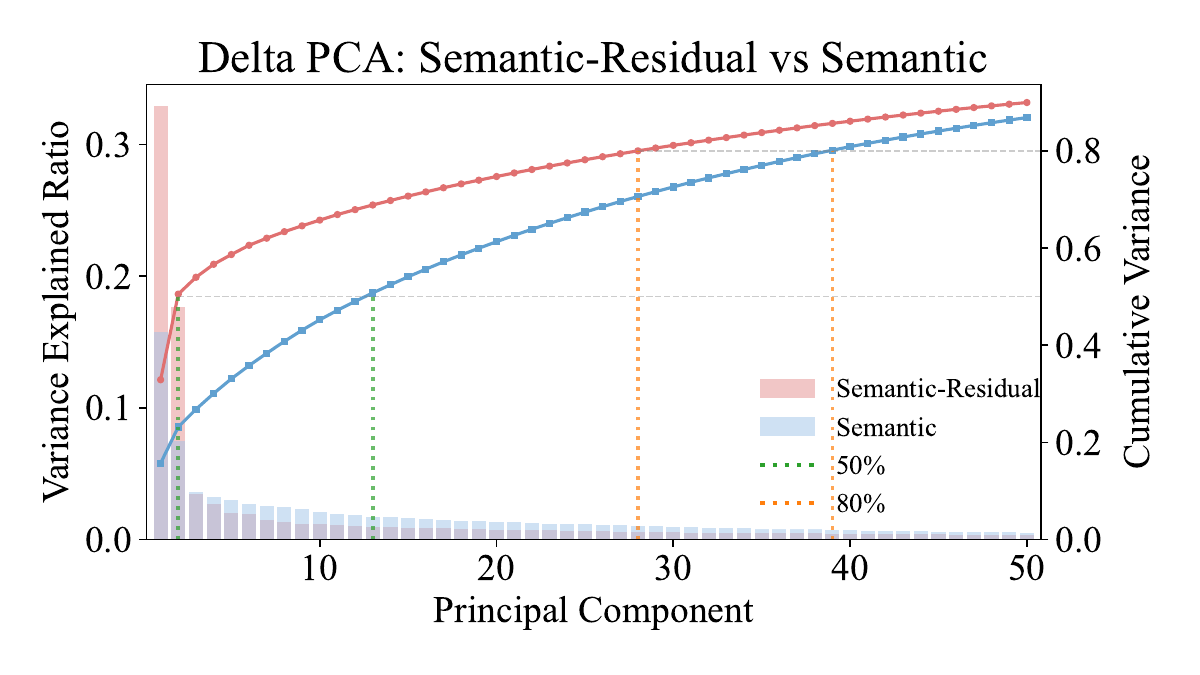}
  \caption{PCA variance spectra of the reconstructed transition deltas ($\Delta$). The semantic-residual subspace exhibits a stronger low-rank tendency. 
  }
  \label{fig:pca}
\end{figure}

\noindent \textbf{Low-Rank Collapse.} 
To further analyze this structural deviation, we perform PCA on the feature deltas ($\Delta$).
The resulting variance spectra (\cref{fig:pca}) reveal a clear low-rank structure largely confined to the semantic-residual subspace.
The first two principal components account for 29.1\% and 16.1\% of the variance, indicating that the structural discrepancy concentrates on a few dominant directions. In contrast, the semantic subspace exhibits a more dispersed variance distribution (PC1: 10.7\%, PC2: 7.4\%), suggesting that macroscopic semantic representations are largely unaffected by this geometric compression.

\noindent \textbf{Universality and Robustness.} Having established the low-rank collapse as a definitive structural signature of the SD1.5 VAE bottleneck, two critical questions naturally arise regarding its viability for real-world forgery detection: 
\begin{itemize}
    \item \textbf{Cross-Architectural Universality:} Is this pronounced geometric deformation an isolated artifact unique to the SD1.5 decoding stage, or a widespread flaw shared across diverse generative paradigms?
    \item \textbf{Robustness against Degradations:} Can this intrinsic manifold deviation withstand superficial pixel-level perturbations, such as severe image compression?
\end{itemize}

\begin{figure}[tb]
  \centering
  \includegraphics[width=\linewidth]{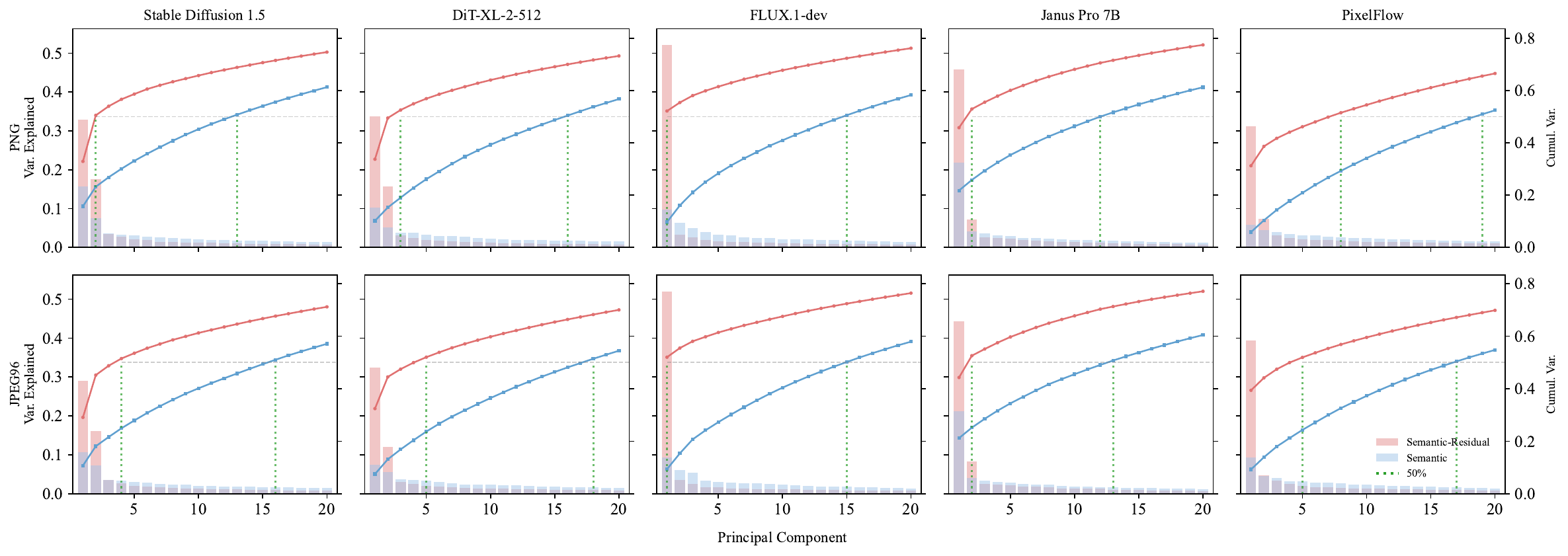}
  \caption{
    {\textbf{PCA variance spectra of counterpart deltas ($\Delta$).} Comparison of uncompressed PNG (\textbf{top}) and JPEG (Q=96) (\textbf{bottom}) reconstructions across diverse pipelines. The consistent eigenvalue distributions verify the universality and robustness of the semantic-residual structural deviation.}
  }
  \label{fig:pca_matrix}
\end{figure}

\noindent Regarding cross-architectural universality, we replicate this exact orthogonal decoupling analysis on mainstream architectures representing fundamentally different generative paradigms, including Stable Diffusion V1.5\cite{rombach2022high}, DiT-XL-2-512\cite{peebles2023scalable}, FLUX.1-dev\cite{flux1dev2024}, PixelFlow\cite{chen2025pixelflow} and JanusPro-7B\cite{chen2025janus}. As shown in \cref{fig:pca_matrix}, the empirical results are remarkably consistent: regardless of their distinct underlying mechanisms, every model exhibits a similarly pronounced low-rank tendency within the semantic-residual subspace.

Similarly, we apply JPEG compression (quality factor 96) to the reconstructed images as a preprocessing step, and then repeat the same analysis pipeline. The core conclusions remain unchanged (\cref{fig:pca_matrix}): the pronounced low-rank tendency in the semantic-residual subspace persists under this lossy perturbation. This indicates that the observed structural deviation is not a trivial artifact of high-frequency reconstruction details, but rather reflects an intrinsic, stable geometric property of the decoding process.

\begin{figure}[tb]
  \centering
  \includegraphics[width=\linewidth]{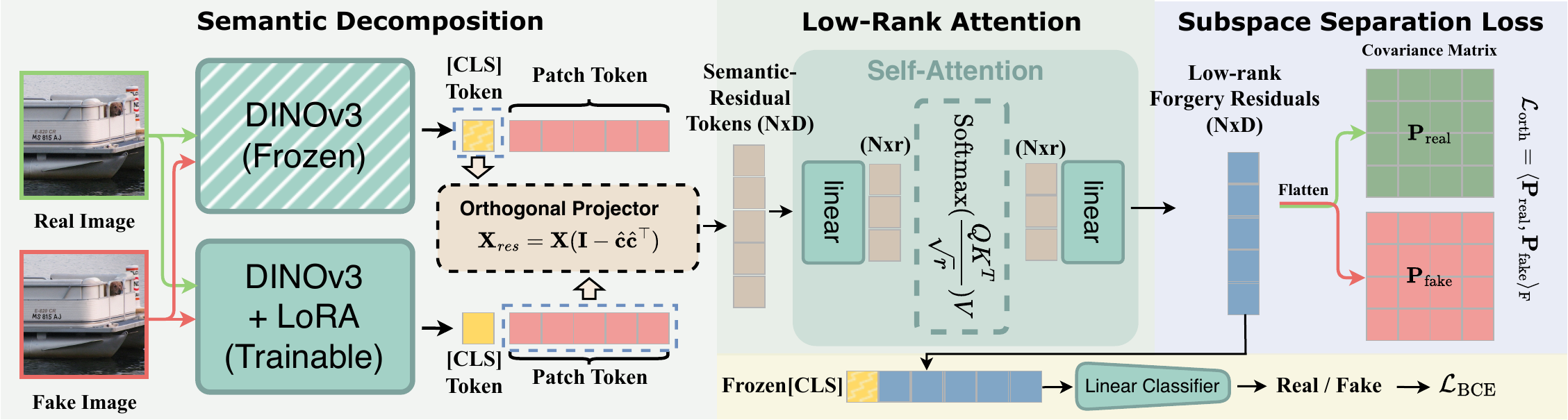}
  \caption{
    LoRC Architecture. DINOv3 patch features are decomposed using the normalized frozen \texttt{[CLS]} token as a semantic anchor, projecting tokens onto its orthogonal complement to obtain semantic-residuals. A Low-Rank Attention Block performs self-attention in a rank-$r$ bottleneck to amplify structured low-rank forensic cues. The pooled residual feature, concatenated with the frozen \texttt{[CLS]} token, is fed to a linear classifier. An auxiliary Subspace Separation Loss further encourages decorrelation between real and fake residual subspaces, reinforcing their geometric distinction.
  }
  \label{fig:method}
\end{figure}

\subsection{Method}
The above analyses consistently reveal a pronounced low-rank structure in the semantic-residual subspace, which remains stable across architectures and robust under lossy perturbations such as JPEG compression.
This suggests that the phenomenon is intrinsic to the generative mechanisms rather than an artifact of specific models.
Motivated by these insights, we propose \textbf{LoRC} (leveraging \textbf{Lo}w-\textbf{R}ank \textbf{C}ollapse AIGI Detector), a framework designed to explicitly capture and isolate the low-rank structural artifacts intrinsic to generative models (\cref{fig:method}). 

\noindent\textbf{Semantic Decomposition.} 
We instantiate the geometric operator in Eq.~(\ref{eq:res}) using a \textbf{frozen} DINOv3 encoder. Trained on large-scale natural images, DINOv3 provides a semantically well-aligned feature space, injecting strong semantic priors without task-specific supervision.
Within this architecture, the \texttt{[CLS]} token aggregates global context from all spatial patches and thus provides an effective reference direction for content alignment in the feature space.
To preserve the pretrained natural-manifold geometry, we keep the encoder \textbf{frozen} and normalize the \texttt{[CLS]} token as a sample-specific \textit{semantic anchor} for geometric decoupling.
We then project patch tokens onto the orthogonal complement of this anchor to attenuate dominant semantic components, thereby shifting real/fake discrimination from a generic semantic feature space into a dynamic forgery subspace conditioned on the underlying semantics.

\noindent\textbf{Low-Rank Forensic Signal Modeling.} 
The low-rank collapse induced by generative decoders implies that residual discrepancies concentrate in a structured low-dimensional subspace. However, modern VFM embeddings remain inherently high-dimensional ($D$), which can obscure these structured low-rank patterns among irrelevant feature fluctuations.
To enhance sensitivity to low-rank residual cues, we introduce a \textbf{Low-Rank Attention Block} that performs self-attention in a rank-$r$ bottleneck space ($r \ll D$). Specifically, we project $\mathbf{X}_{res}$ into $\mathbb{R}^r$ to form the Query, Key, and Value embeddings:
\begin{equation}
    \mathbf{Q} = \mathbf{X}_{res}\mathbf{W}^Q, \quad \mathbf{K} = \mathbf{X}_{res}\mathbf{W}^K, \quad \mathbf{V} = \mathbf{X}_{res}\mathbf{W}^V,
\end{equation}
where $\mathbf{W}^{\{Q,K,V\}} \in \mathbb{R}^{D \times r}$. Self-attention is then performed within this compressed rank-$r$ manifold to capture spatial coherence without noise interference:
\begin{equation}
    \mathbf{A} = \text{Softmax}\left(\frac{\mathbf{Q}\mathbf{K}^\top}{\sqrt{r}}\right)\mathbf{V}.
\end{equation}
Finally, the aggregated low-rank features $\mathbf{A} \in \mathbb{R}^{N \times r}$ are projected back to the original dimension via an output linear transformation $\mathbf{W}^O \in \mathbb{R}^{r \times D}$:
\begin{equation}
    \mathbf{Y} = \mathbf{A}\mathbf{W}^O.
\end{equation}
This design ensures that the model focuses exclusively on the dominant low-rank components where the real/fake discrepancy is most pronounced.
Meanwhile, we retain the \emph{frozen} \texttt{[CLS]} token as a global context anchor to condition the decision and concatenate it with the pooled low-rank residual feature before a linear classifier.

\noindent\textbf{Subspace Separation Loss (SSL).}
Binary classification alone does not explicitly enforce the geometric separation we hypothesize: real residuals should span a high-dimensional subspace while fake residuals collapse onto a low-dimensional one. We therefore introduce an auxiliary loss that directly operates on the subspace geometry.
Given a mini-batch, we partition $\mathbf{X}_{res}$ into real and fake subsets according to the labels.
For each subset, all patch residuals are flattened into a matrix $\mathbf{R} \in \mathbb{R}^{BN \times D}$.
We then compute a normalized covariance descriptor:
\begin{equation}
\mathbf{P} = \frac{\mathbf{R}^\top \mathbf{R}}{\|\mathbf{R}^\top \mathbf{R}\|_F}.
\end{equation}
Let $\mathbf{P}_{\text{real}}$ and $\mathbf{P}_{\text{fake}}$ denote the descriptors of real and fake residuals.
To encourage their subspaces to diverge, we minimize their Frobenius inner product:
\begin{equation}
\mathcal{L}_{\text{SS}} =
\langle \mathbf{P}_{\text{real}}, \mathbf{P}_{\text{fake}} \rangle_F .
\end{equation}
The total training objective is $\mathcal{L} = \mathcal{L}_{\text{BCE}} + \lambda_{\text{SS}}\, \mathcal{L}_{\text{SS}}$.

\section{Experiments}

\subsection{Implementation}
We employ DINOv3 ViT-H+/16~\cite{simeoni2025dinov3} as the backbone, fine-tuned via LoRA~\cite{hu2022lora} with rank 16 and scaling factor $\alpha = 16$. The model is trained on the DDA-Training-Set, where fake samples are generated via SD2.1\cite{rombach2022high} VAE reconstruction of MSCOCO\cite{coco} source images. This setup aligns semantic content, compelling the model to focus specifically on capturing low-rank collapse effects. Images are randomly cropped to $224 \times 224$ for training and center-cropped for inference, with padding applied for insufficient dimensions. We use the Adam~\cite{kingma2015adam} optimizer ($lr=10^{-4}$) with a batch size of $64$ and $4$ gradient accumulation steps. The Low-Rank Attention Block rank is set to $r=32$, and $\lambda_{\text{SS}}=0.1$. 

\subsection{Evaluation Protocol and Benchmark}
We evaluate the out-of-distribution (OOD) generalization capabilities across a series of benchmarks. Results for all baseline methods are derived from their official implementations and pre-trained weights. All testing follows the protocol of DDA~\cite{chen2025dual}. PNG images are recompressed using JPEG (quality factor 96) for consistency. Balanced Accuracy (B.Acc) is used as the primary evaluation metric.

\begin{table}[tpb!]
\centering
\caption{\textbf{Overall comparison across 7 benchmarks.} To ensure fairness and reproducibility, we use official checkpoints released by each method. JPEG compression with a quality factor of 96 is applied to the synthetic images in GenImage and AIGCDetectionBenchmark to mitigate format bias. Bold numbers indicate the best performance.}
\label{tab:compare-methods}
\begin{adjustbox}{width=1.0\linewidth}
\begin{tabular}{lcccccccc}
\toprule
\multirow{2}{*}{Method} & \multicolumn{4}{c}{Standard} & \multicolumn{2}{c}{In-the-Wild} & \multicolumn{1}{c}{Recent Generators} & \multirow{2}{*}{Avg.} \\
\cmidrule(lr){2-5} \cmidrule(lr){6-7} \cmidrule(lr){8-8}
 & GenImage & DRCT-2M & Synthbuster & \makecell{AIGCDetection \\ Benchmark} & Chameleon & WildRF & T2I-CoReBench \\
\midrule
NPR \textsubscript{\textcolor{blue}{(CVPR'24)}} \cite{tan2024rethinking} & 51.5 & 37.3 & 50.0 & 53.1 & 59.9 & 63.5 & 60.0 & 53.6 \\
UnivFD \textsubscript{\textcolor{blue}{(CVPR'23)}} \cite{ojha2023towards} & 64.1 & 61.8 & 67.8 & 72.5 & 50.7 & 55.3 & 10.8 & 54.7 \\
FatFormer \textsubscript{\textcolor{blue}{(CVPR'24)}} \cite{liu2024forgery} & 62.8 & 52.2 & 56.1 & 85.0 & 51.2 & 58.9 & 7.7 & 53.4 \\
SAFE \textsubscript{\textcolor{blue}{(KDD'25)}} \cite{li2025improving} & 50.3 & 59.3 & 46.5 & 50.3 & 59.2 & 57.2 & 1.0 & 46.3 \\
C2P-CLIP \textsubscript{\textcolor{blue}{(AAAI'25)}} \cite{tan2025c2p} & 74.4 & 59.2 & 68.5 & 81.4 & 51.1 & 59.6 & 70.3 & 66.4 \\
AIDE \textsubscript{\textcolor{blue}{(ICLR'25)}} \cite{yan2025sanity} & 61.2 & 64.6 & 53.9 & 63.6 & 63.1 & 58.4 & 9.5 & 53.5 \\
DRCT \textsubscript{\textcolor{blue}{(ICML'24)}} \cite{chen2024drct} & 84.7 & 90.5 & 81.3 & 81.4 & 56.6 & 50.6 & 48.6 & 71.0 \\
AlignedForensics \textsubscript{\textcolor{blue}{(ICLR'25)}} \cite{rajan2025aligned} & 79.0 & 95.5 & 77.4 & 66.6 & 71.0 & 80.1 & 42.6 & 73.2 \\
DDA \textsubscript{\textcolor{blue}{(NIPS'25)}}\cite{chen2025dual} & 91.7 & 98.1 & 90.1 & 87.8 & 82.4 & 90.3 & 91.1 & 90.2 \\
\midrule
\textbf{LoRC (ours)} & \textbf{97.9} & \textbf{99.3} & \textbf{99.9} & \textbf{96.3} & \textbf{92.6} & \textbf{97.6} & \textbf{97.0} & \textbf{97.2} \\
\bottomrule
\end{tabular}
\end{adjustbox}
\end{table}

We categorize the evaluation benchmarks into three groups:
(1) \textbf{Standard Benchmarks.} 
GenImage\cite{zhu2023genimage}, AIGCDetectionBenchmark\cite{zhong2024patchcraft}, Synthbuster\cite{bammey2023synthbuster}, DRCT-2M\cite{chen2024drct}. 
(2) \textbf{In-the-wild Benchmarks.} To evaluate performance against unpredictable real-world degradations and unconstrained generation pipelines, we test on complex in-the-wild datasets, such as Chameleon\cite{yan2025sanity} and WildRF\cite{cavia2024real}.
(3) \textbf{Recent SOTA Generators.} To confront the rapid evolution of generative technologies, we perform extensive zero-shot evaluation on \textbf{T2I-CoReBench}~\cite{li2026easier}. This benchmark comprises images generated from 1,080 carefully curated challenging prompts, deliberately designed to cover complex compositional structures and reasoning-intensive scenarios under real-world conditions. 
It includes \textbf{39} cutting-edge text-to-image (T2I) models spanning diverse architectures, including \textbf{Diffusion Models} (e.g., SD-3.5-Large\cite{esser2024scaling}, FLUX.2-klein-9B~\cite{flux2klein9b2026}, Z-Image-Turbo~\cite{team2025zimage}), \textbf{Autoregressive Models} (e.g., Infinity-8B~\cite{han2025infinity}, GoT-R1-7B\cite{duan2025got}), \textbf{Unified Models} (e.g., Show-o2-7B~\cite{xie2025show}, OmniGen2-7B\cite{wu2025omnigen2}, Hunyuan\-Image-3.0~\cite{cao2025hunyuanimage}), as well as \textbf{closed-source commercial models} (e.g., Seedream 4.5~\cite{seedream4_5} and Nano-Banana-Pro~\cite{nanobananapro}). Each model contributes a substantial volume of \textbf{4,320} generated images. Overall, T2I-CoReBench supports large-scale benchmarking for AIGI detection across diverse generated model families.

\subsection{Overall Results}
As summarized in \cref{tab:compare-methods}, LoRC performs consistently well across all seven datasets under three evaluation settings, achieving an overall accuracy of \textbf{97.2\%} and surpassing the second-best method by \textbf{7.0\%}. This consistent advantage indicates that LoRC generalizes well across diverse data distributions.
On standard benchmarks, LoRC delivers clear and consistent improvements over prior methods on GenImage, DRCT-2M, Synthbuster and AIGCDetectionBenchmark. In particular, it attains a near-perfect mean accuracy on Synthbuster (\cref{tab:compare-synthbuster}), suggesting strong generalization beyond source-specific artifacts.
\textbf{On in-the-wild datasets}, LoRC achieves 92.6\% accuracy on Chameleon and 97.6\% on WildRF. Notably, on the challenging WildRF benchmark, LoRC outperforms the competitive DDA by \textbf{7.3\%}, highlighting its effectiveness in highly diverse and uncurated real-world scenarios. Detailed results for each benchmark are provided in the Supplementary Material.

\subsection{Zero-Shot Comparison on SOTA Generators.}
As generative models evolve at an unprecedented pace, generalizing to unseen architectures remains the most formidable challenge in AIGI detection. We evaluate this zero-shot capability on T2I-CoReBench (\cref{tab:t2i-corebench}). Interestingly, while most baseline methods suffer severe performance degradation, only DDA maintains moderate robustness by aligning frequency-domain statistics. In contrast, by explicitly capturing the universal low-rank collapse injected by the final decoding stage, our proposed LoRC achieves a state-of-the-art average accuracy of \textbf{97.0\%}, yielding an impressive improvement of \textbf{5.9\%} over existing approaches. These compelling results thoroughly validate the unparalleled robustness of our method, highlighting its immense potential for defending against rapidly emerging generative models.

\begin{table}[tb!]
\centering
\caption{Comparison of balanced accuracy on Synthbuster.}
\label{tab:compare-synthbuster}
\begin{adjustbox}{width=\linewidth}
\begin{tabular}{lcccccccccl}
\toprule
Method & DALL·E 2 & DALL·E 3 & Firefly & GLIDE & Midjourney & SD 1.3 & SD 1.4 & SD 2 & SDXL & Avg. \\
\midrule
NPR \textsubscript{\textcolor{blue}{(CVPR'24)}} \cite{tan2024rethinking} 
 & 51.1 & 49.3 & 46.5 & 48.5 & 52.8 & 51.4 & 51.8 & 46.0 & 52.8 & 50.0 \\
UnivFD \textsubscript{\textcolor{blue}{(CVPR'23)}} \cite{ojha2023towards} 
& 83.5 & 47.4 & 89.9 & 53.3 & 52.5 & 70.4 & 69.9 & 75.7 & 68.0 & 67.8 \\
FatFormer \textsubscript{\textcolor{blue}{(CVPR'24)}} \cite{liu2024forgery} 
& 59.4 & 39.5 & 60.3 & 72.7 & 44.4 & 53.7 & 54.0 & 52.3 & 69.1 & 56.1 \\
SAFE \textsubscript{\textcolor{blue}{(KDD'25)}} \cite{li2025improving} 
& 58.0 & 9.9 & 10.3 & 52.2 & 56.7 & 59.4 & 59.1 & 53.0 & 59.5 & 46.5 \\
C2P-CLIP \textsubscript{\textcolor{blue}{(AAAI'25)}} \cite{tan2025c2p} 
& 55.6 & 63.2 & 59.5 & \underline{86.7} & 52.9 & 75.2 & 76.7 & 69.2 & 77.7 & 68.5 \\
AIDE \textsubscript{\textcolor{blue}{(ICLR'25)}} \cite{yan2025sanity} 
& 34.9 & 33.7 & 24.8 & 65.0 & 57.5 & 74.1 & 73.7 & 53.2 & 68.4 & 53.9 \\
DRCT \textsubscript{\textcolor{blue}{(ICML'24)}} \cite{chen2024drct} 
& 77.2 & 86.6 & 84.1 & 82.6 & 73.7 & 86.6 & 86.6 & 83.2 & 71.3 & 81.3 \\
AlignedForensics \textsubscript{\textcolor{blue}{(ICLR'25)}} \cite{rajan2025aligned} 
& 50.2 & 48.9 & 51.7 & 53.5 & \underline{98.7} & \underline{98.8} & \underline{98.8} & \underline{98.6} & \underline{97.3} & 77.4 \\
DDA \textsubscript{\textcolor{blue}{(NIPS'25)}}\cite{chen2025dual}
& \underline{86.3} & \underline{90.0} & \underline{91.9} & 76.5 & 93.5 & 92.9 & 92.7 & 93.3 & 93.5 & \underline{90.1} \\
\midrule
\textbf{LoRC (ours)} 
& \textbf{99.6} & \textbf{99.9} & \textbf{99.6} & \textbf{99.8} & \textbf{100.0} & \textbf{100.0} & \textbf{100.0} & \textbf{100.0} & \textbf{100.0} & \textbf{99.9} \\
\bottomrule
\end{tabular}
\end{adjustbox}
\end{table}

\begin{table}[t]
\centering
\small
\setlength{\tabcolsep}{5pt}
\renewcommand{\arraystretch}{1.1}
\caption{Ablation study of the LoRC where modules are added cumulatively.}
\resizebox{0.9\linewidth}{!}{
\begin{tabular}{lcccc}
\toprule
\textbf{Setting} & \textbf{Standard} & \textbf{In-the-wild} & \textbf{T2I-CoReBench} & \textbf{Avg.} \\
\midrule
Baseline & 97.9 & 89.4 & 94.0 & 93.8 \\
+ Semantic Decomposition & 97.2 & 86.9 & \textbf{98.1} & 94.1 \\
+ Low-Rank Attention & 96.9 & 90.2 & 96.0 & 94.4 \\
+ Subspace Separation Loss & \textbf{98.1} & \textbf{95.2} & 97.0 & \textbf{96.8} \\
\bottomrule
\end{tabular}
}
\label{tab:ablation}
\end{table}

\begin{table*}[!t]
\setlength{\tabcolsep}{3pt}
\centering
\caption{Comparison of fake accuracy on T2I-CoReBench. C2P., Fat., and Align. are abbreviations for C2P-CLIP, FatFormer, and AlignedForensics, respectively. All test images are compressed using JPEG with a quality factor of 96.}
\label{tab:t2i-corebench}
\resizebox{\linewidth}{!}{
    \begin{tabular}{l cccccccccc}
        \toprule
        \textbf{Generator} & \textbf{NPR} & \textbf{UnivFD} & \textbf{C2P.} & \textbf{Fat.} & \textbf{SAFE} & \textbf{AIDE} & \textbf{DRCT} & \textbf{Align.} & \textbf{DDA} & \cellcolor{blue!5}\textbf{LoRC} \\
        \midrule
        \multicolumn{11}{c}{\textit{Diffusion Models}} \\
        \midrule
        SD-3-Medium       & 51.6 & 5.0 & 71.0 & 3.8 & 1.6 & 19.2 & 59.4 & 60.5 & 96.8 & \cellcolor{blue!5}\textbf{99.6} \\
        SD-3.5-Medium     & 61.6 & 9.9 & 66.7 & 12.9 & 1.6 & 7.2 & 42.4 & 68.5 & 97.0 & \cellcolor{blue!5}\textbf{99.7} \\
        SD-3.5-Large      & 66.1 & 12.7 & 53.0 & 7.6 & 1.4 & 4.2 & 57.0 & 58.9 & 97.0 & \cellcolor{blue!5}\textbf{99.8} \\
        FLUX.1-schnell    & 51.6 & 4.5 & 59.9 & 2.5 & 0.7 & 16.8 & 36.0 & 27.1 & 94.8 & \cellcolor{blue!5}\textbf{98.0} \\
        FLUX.1-dev        & 58.7 & 3.5 & 73.5 & 0.8 & 0.4 & 18.7 & 33.5 & 21.0 & 90.6 & \cellcolor{blue!5}\textbf{96.2} \\
        FLUX.1-Krea-dev   & 44.9 & 4.6 & \textbf{86.4} & 1.0 & 2.6 & 15.1 & 22.6 & 4.2 & 30.7 & \cellcolor{blue!5}81.2 \\
        FLUX.2-dev        & 65.9 & 3.0 & 75.9 & 1.7 & 0.6 & 4.0 & 15.8 & 2.0 & 82.7 & \cellcolor{blue!5}\textbf{91.2} \\
        FLUX.2-klein-4B   & 52.7 & 3.2 & 75.0 & 1.3 & 0.5 & 3.6 & 12.3 & 4.1 & 90.5 & \cellcolor{blue!5}\textbf{96.6} \\
        FLUX.2-klein-9B   & 57.0 & 1.7 & 67.9 & 0.8 & 0.5 & 2.1 & 4.6 & 1.7 & 91.0 & \cellcolor{blue!5}\textbf{94.9} \\
        PixArt-$\alpha$   & 68.2 & 16.0 & 72.9 & 16.3 & 1.5 & 20.8 & 79.8 & 96.2 & 99.5 & \cellcolor{blue!5}\textbf{99.8} \\
        PixArt-$\Sigma$   & 68.5 & 15.1 & 67.6 & 19.3 & 1.9 & 23.3 & 70.3 & 89.5 & 99.5 & \cellcolor{blue!5}\textbf{99.9} \\
        HiDream-I1        & 55.4 & 2.6 & 60.0 & 0.3 & 0.4 & 9.1 & 54.9 & 33.5 & 83.6 & \cellcolor{blue!5}\textbf{97.7} \\
        Qwen-Image        & 86.2 & 5.2 & 93.6 & 1.8 & 0.3 & 10.8 & 35.7 & 21.7 & 97.8 & \cellcolor{blue!5}\textbf{98.6} \\
        Qwen-Image-2512   & 93.6 & 9.2 & 94.4 & 3.5 & 0.1 & 1.8 & 27.6 & 86.2 & 98.7 & \cellcolor{blue!5}\textbf{99.6} \\
        LongCat-Image     & 69.1 & 2.9 & 65.1 & 1.8 & 2.0 & 2.7 & 14.6 & 2.1 & 95.0 & \cellcolor{blue!5}\textbf{99.7} \\
        Z-Image           & 65.6 & 6.1 & 74.9 & 4.5 & 0.6 & 9.5 & 54.2 & 13.7 & 96.9 & \cellcolor{blue!5}\textbf{98.7} \\
        Z-Image-Turbo     & 43.3 & 3.5 & 72.9 & 0.4 & 0.4 & 10.4 & 56.9 & 3.2 & 76.4 & \cellcolor{blue!5} \textbf{92.8} \\
        \midrule
        \multicolumn{11}{c}{\textit{Autoregressive Models}} \\
        \midrule
        Infinity-8B       & 67.5 & 26.9 & 80.4 & 24.3 & 0.7 & 4.3 & 78.7 & 80.0 & 99.8 & \cellcolor{blue!5}\textbf{100.0} \\
        GoT-R1-7B         & 7.7 & 16.0 & 72.3 & 12.8 & 0.4 & 6.0 & 42.1 & 38.4 & 100.0 & \cellcolor{blue!5}\textbf{100.0} \\
        \midrule
        \multicolumn{11}{c}{\textit{Unified Models}} \\
        \midrule
        BAGEL             & 56.4 & 8.9 & 63.1 & 2.6 & 0.7 & 14.6 & 55.3 & 17.1 & 95.4 & \cellcolor{blue!5}\textbf{99.8} \\
        BAGEL w/Think     & 62.3 & 7.1 & 74.9 & 2.9 & 0.6 & 19.2 & 57.0 & 16.3 & 96.9 & \cellcolor{blue!5}\textbf{100.0} \\
        show-o2-1.5B      & 60.5 & 16.5 & 73.2 & 15.1 & 0.3 & 20.1 & 83.1 & 81.3 & 99.4 & \cellcolor{blue!5}\textbf{100.0} \\
        show-o2-7B        & 9.9 & 10.0 & 53.8 & 15.4 & 0.1 & 3.0 & 61.6 & 31.9 & 99.7 & \cellcolor{blue!5}\textbf{100.0} \\
        Janus-Pro-1B      & 11.3 & 44.9 & 71.6 & 38.8 & 0.7 & 10.0 & 52.1 & 54.6 & 99.4 & \cellcolor{blue!5}\textbf{99.9} \\
        Janus-Pro-7B      & 12.9 & 25.2 & 68.3 & 16.1 & 0.4 & 7.2 & 52.3 & 59.0 & 99.7 & \cellcolor{blue!5}\textbf{100.0} \\
        BLIP3o-4B         & 52.9 & 19.9 & 65.6 & 29.8 & 2.2 & 6.0 & 67.5 & 75.5 & 99.7 & \cellcolor{blue!5}\textbf{100.0} \\
        BLIP3o-8B         & 49.8 & 24.2 & 56.4 & 25.3 & 2.1 & 8.9 & 63.7 & 78.5 & 99.6 & \cellcolor{blue!5}\textbf{99.9} \\
        OmniGen2-7B       & 58.3 & 8.6 & 77.7 & 7.9 & 1.3 & 28.5 & 55.1 & 37.9 & 97.0 & \cellcolor{blue!5}\textbf{99.6} \\
        HunyuanImage-3.0  & 91.8 & 9.5 & 77.5 & 3.4 & 3.6 & 3.5 & 72.3 & 99.0 & 99.3 & \cellcolor{blue!5}\textbf{99.9} \\
        \midrule
        \multicolumn{11}{c}{\textit{Closed-Source Models}} \\
        \midrule
        Seedream 3.0      & 56.0 & 4.1 & 80.6 & 0.2 & 0.2 & 20.2 & 52.8 & 19.1 & 95.8 & \cellcolor{blue!5}\textbf{99.5} \\
        Seedream 4.0      & 90.4 & 22.8 & 58.4 & 5.5 & 0.7 & 0.0 & 77.5 & 81.0 & 95.4 & \cellcolor{blue!5}\textbf{99.9} \\
        Seedream 4.5      & 92.6 & 26.2 & 67.0 & 4.2 & 0.3 & 0.0 & 45.5 & 80.6 & 95.7 & \cellcolor{blue!5}\textbf{99.6} \\
        Gemini 2.0 Flash  & 49.9 & 15.7 & 74.8 & 4.1 & 2.3 & 23.8 & 73.3 & 23.8 & 94.2 & \cellcolor{blue!5}\textbf{98.5} \\
        Nano Banana       & 67.5 & 3.8 & 54.8 & 2.4 & 0.7 & 1.6 & 50.9 & 43.4 & 70.2 & \cellcolor{blue!5} \textbf{94.8} \\
        Nano Banana Pro   & 79.5 & 4.4 & 86.0 & 2.2 & 0.5 & 2.9 & 80.7 & 80.7 & 93.2 & \cellcolor{blue!5} \textbf{94.8} \\
        Imagen 4          & 57.4 & 2.9 & 74.2 & 1.6 & 0.1 & 6.0 & 45.5 & 27.3 & \textbf{90.7} & \cellcolor{blue!5}85.5 \\
        Imagen 4 Ultra    & 73.9 & 3.4 & 75.1 & 1.7 & 0.2 & 5.3 & 48.0 & 34.9 & \textbf{90.4} & \cellcolor{blue!5}88.2 \\
        GPT-Image 1.5     & 82.9 & 5.4 & 75.4 & 1.4 & 4.1 & 0.3 & 2.4 & 6.9 & 57.5 & \cellcolor{blue!5}\textbf{91.1} \\
        GPT-4o            & 87.9 & 4.2 & 30.3 & 2.7 & 1.2 & 1.0 & 0.9 & 0.1 & 66.5 & \cellcolor{blue!5}\textbf{88.8} \\
        \midrule
        \textbf{Average}  & 60.0 & 10.8 & 70.3 & 7.7 & 1.0 & 9.5 & 48.6 & 42.6 & \underline{91.1} & \cellcolor{blue!5}\textbf{97.0} \\
        \bottomrule
    \end{tabular}
}
\end{table*}

\subsection{Ablation Study}
Table~\ref{tab:ablation} summarizes an ablation study where we cumulatively add each component of LoRC. 
Semantic Decomposition notably improves T2I-CoReBench\cite{li2026easier}, indicating that suppressing dominant semantics is crucial for modern generators with strong semantic alignment. Low-Rank Attention boosts in-the-wild performance by better modeling the collapsed low-rank residual cues. Finally, Subspace Separation Loss enlarges the real/fake subspace gap, yielding the best overall results.

\subsection{Hyperparameter Sensitivity}
\begin{figure}[tb]
    \centering
    \begin{minipage}{\textwidth} 
        \centering
        \begin{subfigure}[b]{0.32\textwidth}
            \includegraphics[width=\textwidth]{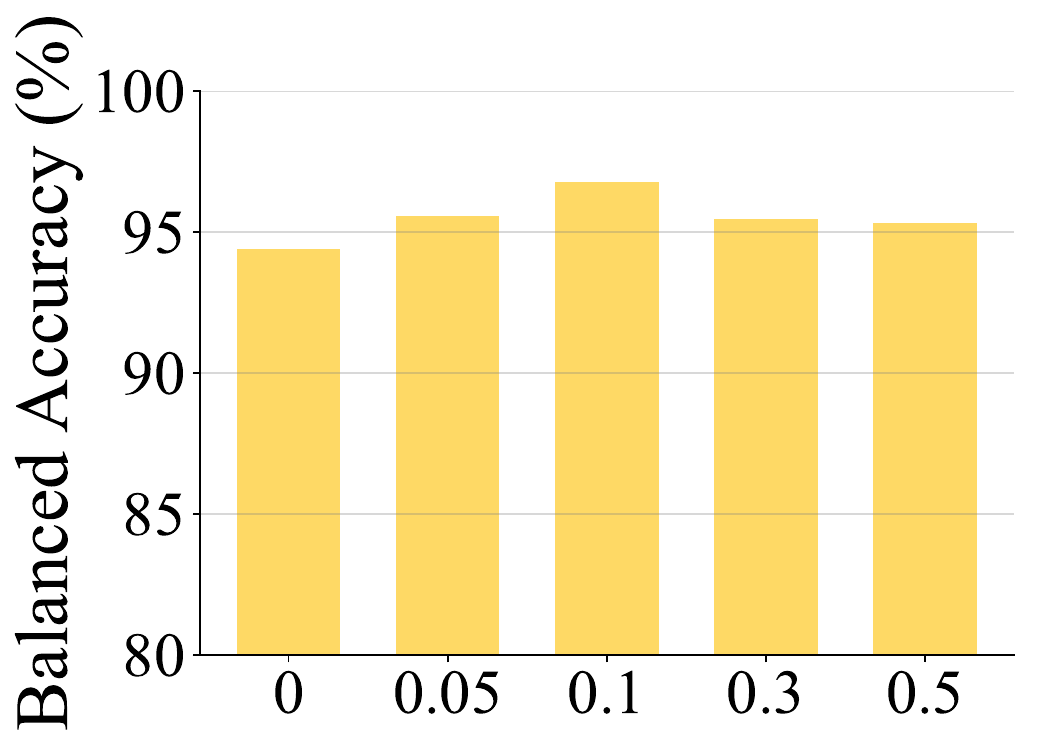}
            \caption{$\lambda_{SS}$}
            \label{fig:ablation-lambda-ss}
        \end{subfigure}%
        \hfill%
        \begin{subfigure}[b]{0.32\textwidth}
            \includegraphics[width=\textwidth]{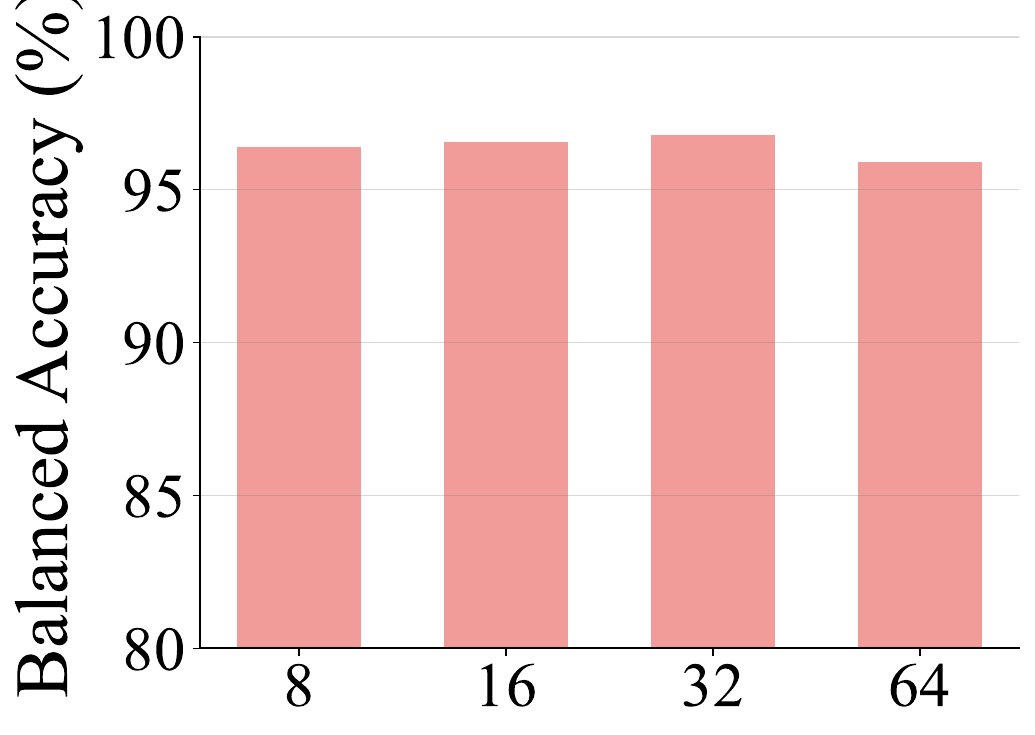}
            \caption{Low-Rank Attention Rank}
            \label{fig:ablation-rank}
        \end{subfigure}%
        \hfill%
        \begin{subfigure}[b]{0.32\textwidth}
            \includegraphics[width=\textwidth]{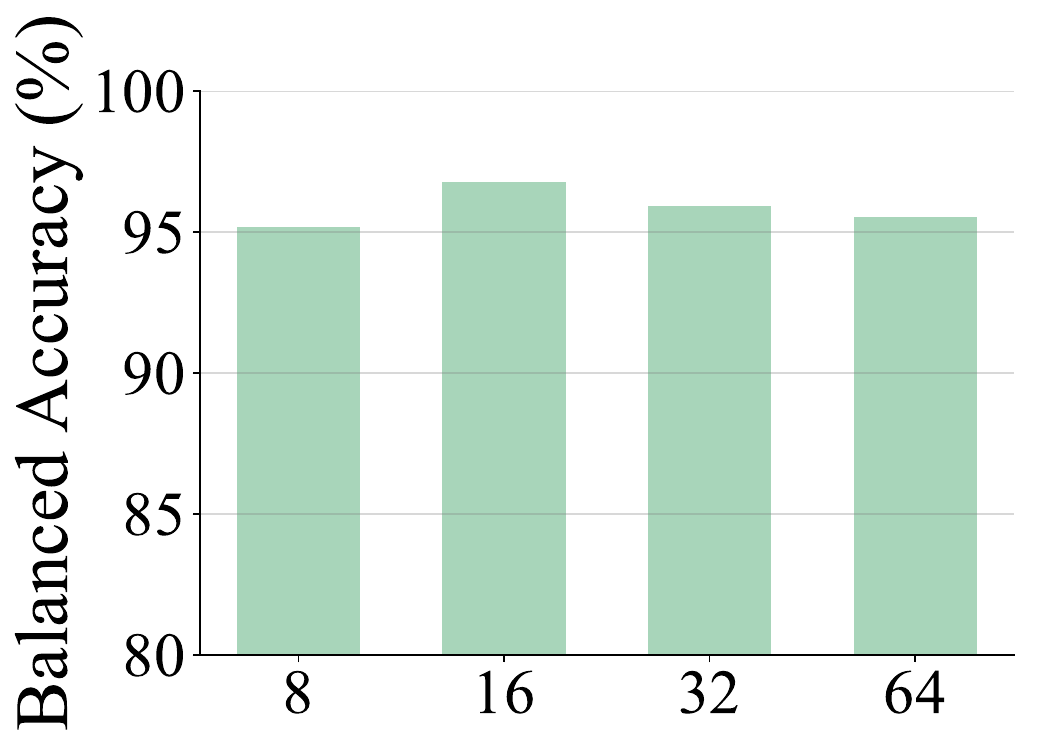}
            \caption{LoRA Rank}
            \label{fig:ablation-lora-rank}
        \end{subfigure}
    \end{minipage}
    \caption{\textbf{Hyperparameter sensitivity.} 
    (a) The impact of the weight parameter $\lambda_{SS}$ on Subspace Separation Loss.
    (b) The impact of the rank in the Low-Rank Attention Block.
    (c) The effect of varying the LoRA rank during fine-tuning.
    }
    \label{fig:ablation-hyper}
\end{figure}

\cref{fig:ablation-hyper} shows the ablation results of three key hyperparameters. 
The subspace separation loss improves performance when $\lambda_{SS}>0$, with the best balanced accuracy at $\lambda_{SS}=0.1$. 
The optimal feature rank for the Low-Rank Attention Block is $32$, indicating a suitable trade-off between compression and information preservation. 
For LoRA, rank $16$ gives the best result by balancing adaptation capacity and parameter efficiency.

\begin{figure}[tb]
    \centering
    \begin{minipage}{\textwidth} 
        \centering
        \includegraphics[width=0.99\textwidth]{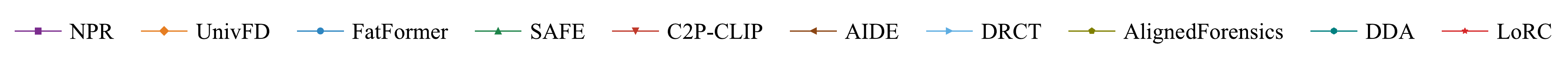}
        \begin{subfigure}[b]{0.32\textwidth}
            \includegraphics[width=\textwidth]{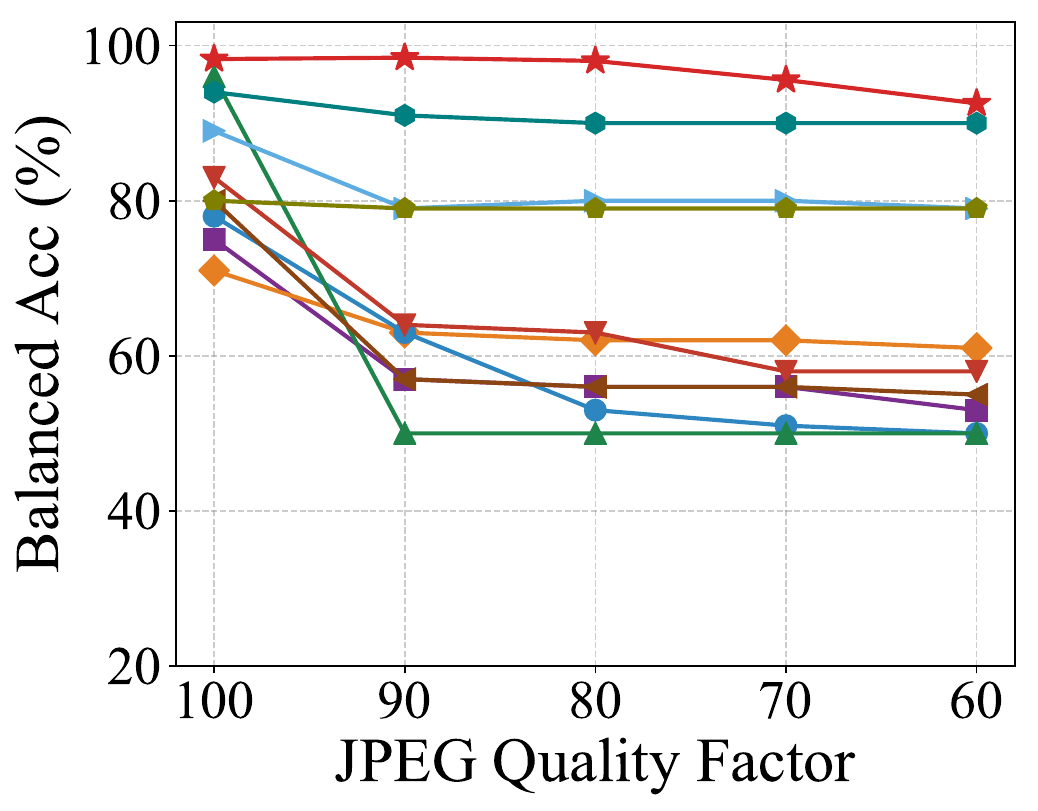}
        \end{subfigure}%
        \hfill%
        \begin{subfigure}[b]{0.32\textwidth}
            \includegraphics[width=\textwidth]{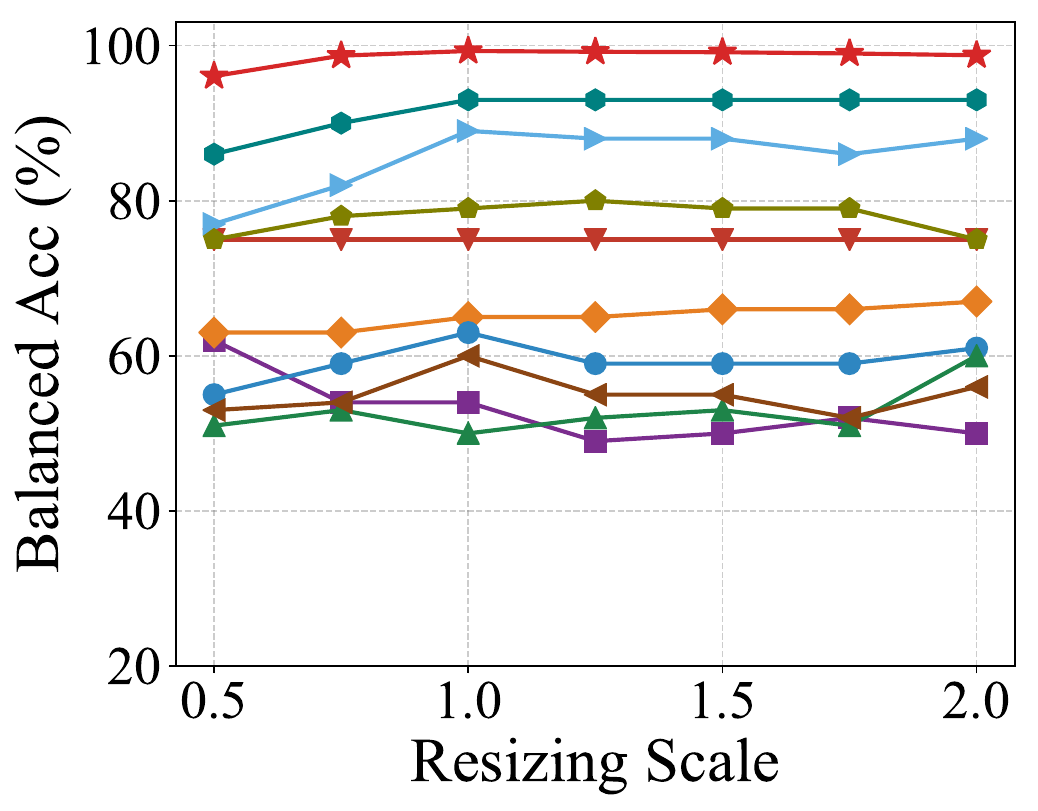}
        \end{subfigure}%
        \hfill%
        \begin{subfigure}[b]{0.32\textwidth}
            \includegraphics[width=\textwidth]{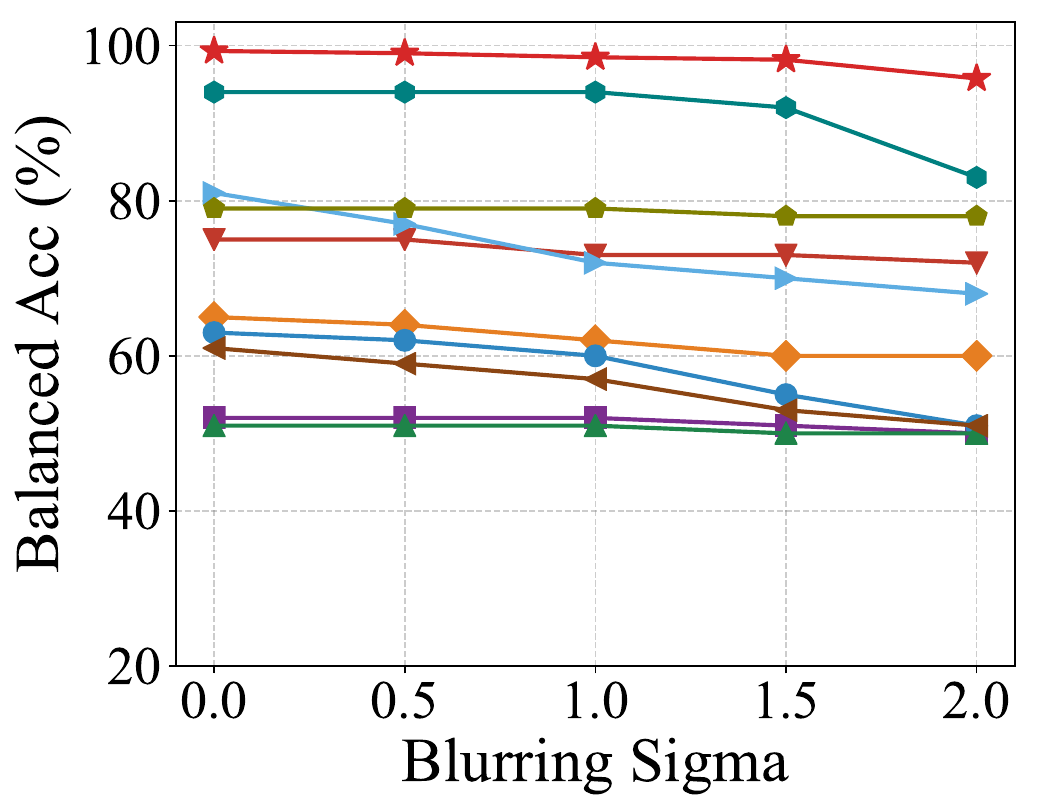}
        \end{subfigure}
    \end{minipage}
    \caption{Robustness analysis on GenImage.}
    \label{fig:robustness}
\end{figure}

\subsection{Robustness}
In real-world applications, images frequently undergo various types of degradation. To evaluate practical reliability, we further investigate the robustness of LoRC. \Cref{fig:robustness} illustrates the model performance under three common perturbations: JPEG compression, image resizing, and Gaussian blurring. LoRC maintains consistently high balanced accuracy across all test conditions. This consistent stability demonstrates that LoRC effectively captures the underlying low-rank structural effect, which provides inherent robustness against common pixel-level degradations.

\section{Conclusion}
We revisit AI-generated image detection from a geometric perspective and uncover an architecture-agnostic signature: modern generators consistently induce a pronounced \emph{low-rank collapse} in the semantic-residual orthogonal subspace, while largely preserving the dominant semantic direction. This reveals a stable geometric degeneracy introduced by the final decoding bottleneck, shared across diverse generative paradigms and resilient to common post-processing.
Motivated by this observation, we propose \textbf{LoRC}, which suppresses semantic dominance via orthogonal decomposition and explicitly models the collapsed residual geometry with a low-rank attention bottleneck, further reinforced by a subspace separation objective.
Extensive evaluations demonstrate that LoRC yields a 7.0\% absolute average improvement across 7 different benchmarks, reaching a 97.2\% overall accuracy. In addition, it exhibits strong zero-shot generalization on rapidly evolving generator suites by achieving 97.0\% accuracy across \textbf{39} unseen generators, which firmly supports low-rank collapse as a reliable universal signature for real-world AIGI detection.

\section*{Acknowledgements}
This work was supported in part by the National Natural Science Foundation of China (Grant Nos. 62302295, 62595733, and 62561160155), the Shanghai Municipal Science and Technology Major Project (Grant No. 2021SHZDZX0102). This work was also supported by Tencent Youtu Lab.


%
%
\bibliographystyle{splncs04}
\bibliography{main}

\clearpage  
\begin{center}
    {\LARGE\bfseries Supplementary Material}
\end{center}

\setcounter{section}{0}
\renewcommand{\thesection}{\Alph{section}}
\renewcommand{\thesubsection}{\thesection.\arabic{subsection}}

\noindent 
This supplementary material provides additional details and experimental results for LoRC. \Cref{sec:compared-methods} briefly summarizes the compared methods used in the main paper. \Cref{sec:detailed} presents detailed benchmark results on GenImage, DRCT-2M, AIGCDetectionBenchmark and WildRF. \Cref{sec:ablation} provides more ablation studies, including semantic decomposition, the placement of the subspace separation loss (SSL), backbone analysis. Finally, \Cref{sec:computational-cost} analyzes the computational cost of LoRC and its accuracy-efficiency balance.

\section{Overview of Compared Methods}
\label{sec:compared-methods}

Below we briefly summarize the baseline methods compared in the main paper.

\paragraph{NPR \cite{tan2024rethinking}} is built on a CNN-based detector and targets artifacts introduced by up-sampling operations in generative networks. It characterizes neighboring pixel relationships as transferable low-level forensic cues, enabling the model to capture generator-agnostic traces in synthetic images.

\paragraph{UnivFD \cite{ojha2023towards}} is built upon pretrained CLIP visual features instead of a fully supervised deepfake detector trained from scratch. In addition to the frozen feature extractor, it learns a linear classifier, which alleviates overfitting to generator-specific artifacts and improves transferability across different generative models.

\paragraph{FatFormer \cite{liu2024forgery}} adopts a vision transformer backbone and equips it with a forgery-aware adaptive module. This module enhances forgery-sensitive representations in both the image and frequency domains. In addition, the method aligns image features with textual prompts through contrastive learning, allowing the pretrained model to better attend to synthetic artifacts.

\paragraph{SAFE \cite{li2025improving}} builds a synthetic image detector from the perspective of image transformations. The detector is trained with a series of strategies, including crop-based preprocessing to preserve forensic cues, augmentations to suppress semantic and color biases, and random masking to strengthen local artifact awareness, thereby improving robustness.

\paragraph{C2P-CLIP \cite{tan2025c2p}} is developed on top of CLIP and improves AIGI detection by injecting category-common semantic prompts during training. Specifically, ClipCap is first used to generate image captions, which are then enriched with shared category-level prompts. The resulting caption-image pairs are used to optimize the image encoder through contrastive learning. During inference, only the adapted image encoder and a linear classifier are used.

\paragraph{AIDE \cite{yan2025sanity}} adopts a hybrid architecture that combines high-level semantic representations with low-level forensic clues. It uses CLIP embeddings to encode semantic content, while low-level features are extracted from extreme-frequency regions selected by frequency analysis and enhanced with handcrafted noise cues. These complementary features are then fused for final classification.

\paragraph{DRCT \cite{chen2024drct}} is a reconstruction-based training framework for diffusion-image detection. It reconstructs real images with diffusion models to generate hard synthetic samples that preserve semantic content while introducing subtle generative artifacts. Contrastive learning is then employed to guide the detector toward these artifact patterns.

\paragraph{AlignedForensics \cite{rajan2025aligned}} emphasizes detector training with aligned data. It constructs aligned real-fake pairs by passing real images through the autoencoder of a latent diffusion model in a single forward pass, without denoising. This design largely removes content and semantic mismatch, forcing the detector to focus mainly on artifacts introduced by the generative decoding process.

\paragraph{B-Free \cite{guillaro2024bias}} proposes a bias-reduced training paradigm for general AI-generated image detection. It creates semantically aligned synthetic counterparts of real images through self-conditioned diffusion reconstruction, so that the discrepancy between the two mainly comes from generation artifacts rather than content bias. The method also introduces content augmentation such as inpainting to improve diversity and reduce spurious correlations.

\paragraph{DDA \cite{chen2025dual}} aligns real-fake pairs in both the pixel and frequency domains to reduce dataset bias. Unlike reconstruction-only alignment, it explicitly mitigates frequency-level mismatch, so that the detector focuses less on spurious content or frequency cues.

\section{Detailed Benchmark Results}
\label{sec:detailed}
In this section, we present detailed results on different subsets of the benchmarks, including GenImage (\cref{tab:compare-genimage}), DRCT-2M (\cref{tab:compare-drct}), and AIGCDetectionBenchmark (\cref{tab:aigc-detection}). LoRC achieves the best average performance on all three benchmarks, with consistent gains over the second-best method. In particular, the margin reach 6.2\% on GenImage, 1.2\% on DRCT-2M, and 8.5\% on AIGCDetectionBenchmark. These results indicate that LoRC delivers more reliable detection across different benchmarks and generator families. Furthermore, as shown in \cref{tab:compare-chameleon-wildrf}, LoRC maintains a strong advantage on the Chameleon and WildRF benchmarks. Specifically, it surpasses the second-best method by 10.2\% on Chameleon and by 7.3\% in average balanced accuracy on WildRF, demonstrating its exceptional robustness in diverse and real-world scenarios.

\begin{table}[tbp!]
  \centering
\caption{Comparison of balanced accuracy on GenImage.}
\label{tab:compare-genimage}
 \begin{adjustbox}{width=\linewidth}
      \begin{tabular}{lccccccccl}
      \toprule
      Method & Midjourney & SDv1.4 & SDv1.5 & ADM & GLIDE & Wukong & VQDM & BigGAN & Avg. \\
      \midrule
NPR \textsubscript{\textcolor{blue}{(CVPR'24)}} \cite{tan2024rethinking} & 53.4 & 55.1 & 55.0 & 43.8 & 41.2 & 57.4 & 48.4 & 57.7 & 51.5 \\
UnivFD \textsubscript{\textcolor{blue}{(CVPR'23)}} \cite{ojha2023towards} & 55.1 & 55.6 & 55.7 & 62.5 & 61.3 & 61.1 & 76.9 & 84.4 & 64.1 \\  
FatFormer \textsubscript{\textcolor{blue}{(CVPR'24)}} \cite{liu2024forgery} & 52.1 & 53.6 & 53.8 & 61.4 & 65.5 & 60.9 & 72.5 & 82.2 & 62.8 \\
SAFE \textsubscript{\textcolor{blue}{(KDD'25)}} \cite{li2025improving} & 49.0 & 49.7 & 49.8 & 49.5 & 53.0 & 50.3 & 50.2 & 50.9 & 50.3 \\
C2P-CLIP \textsubscript{\textcolor{blue}{(AAAI'25)}} \cite{tan2025c2p} & 56.6 & 77.5 & 76.9 & 71.6 & 73.5 & 79.4 & 73.7 & 85.9 & 74.4 \\
AIDE \textsubscript{\textcolor{blue}{(ICLR'25)}} \cite{yan2025sanity} & 58.2 & 77.2 & 77.4 & 50.4 & 54.6 & 70.5 & 50.8 & 50.6 & 61.2 \\
DRCT \textsubscript{\textcolor{blue}{(ICML'24)}} \cite{chen2024drct} & 82.4 & 88.3 & 88.2 & 76.9 & 86.1 & 87.9 & \underline{85.4} & \underline{87.0} & 84.7 \\
AlignedForensics \textsubscript{\textcolor{blue}{(ICLR'25)}} \cite{rajan2025aligned} & \underline{97.5} & \textbf{99.7} & \textbf{99.6} & 52.4 & 57.6 & \textbf{99.6} & 75.0 & 50.6 & 79.0 \\
DDA \textsubscript{\textcolor{blue}{(NIPS'25)}}\cite{chen2025dual} & 95.6 & 98.7 & 98.6 & \underline{89.5} & \underline{89.6} & \underline{98.7} & 76.5 & \underline{86.5} & \underline{91.7} \\
\midrule
\textbf{LoRC (ours)} & \textbf{97.8} & \underline{98.8} & \underline{98.8} & \textbf{98.4} & \textbf{98.5} & \underline{98.7} & \textbf{98.8} & \textbf{93.6} & \textbf{97.9} \\
\bottomrule
      \end{tabular}
      \end{adjustbox}
\end{table}

\begin{table}[tbp!]
  \centering
  \caption{Comparison of balanced accuracy on DRCT-2M.}
  \label{tab:compare-drct}
  \begin{adjustbox}{width=1.0\linewidth}
  \begin{tabular}{lccccccccccccccccl}
  \toprule
  Method & LDM & SDv1.4 & SDv1.5 & SDv2 & SDXL & \makecell{SDXL-\\Refiner} & \makecell{SD-\\Turbo} & \makecell{SDXL-\\Turbo} & \makecell{LCM-\\SDv1.5} & \makecell{LCM-\\SDXL} & \makecell{SDv1-\\Ctrl} & \makecell{SDv2-\\Ctrl} & \makecell{SDXL-\\Ctrl} & \makecell{SDv1-\\DR} & \makecell{SDv2-\\DR} & \makecell{SDXL-\\DR} & Avg. \\
  \midrule
NPR \textsubscript{\textcolor{blue}{(CVPR'24)}} \cite{tan2024rethinking} & 33.0 & 29.1 & 29.0 & 35.1 & 33.2 & 28.4 & 27.9 & 27.9 & 29.4 & 30.2 & 28.4 & 28.3 & 34.7 & 67.9 & 67.4 & 66.1 & 37.3 \\
UnivFD \textsubscript{\textcolor{blue}{(CVPR'23)}} \cite{ojha2023towards} & 85.4 & 56.8 & 56.4 & 58.2 & 63.2 & 55.0 & 56.5 & 53.0 & 54.5 & 65.9 & 68.0 & 65.4 & 75.9 & 64.6 & 56.2 & 53.9 & 61.8 \\
FatFormer \textsubscript{\textcolor{blue}{(CVPR'24)}} \cite{liu2024forgery} & 55.9 & 48.2 & 48.2 & 48.2 & 48.2 & 48.3 & 48.2 & 48.2 & 48.3 & 50.6 & 49.7 & 49.9 & 59.8 & 66.3 & 60.6 & 56.0 & 52.2  \\
SAFE \textsubscript{\textcolor{blue}{(KDD'25)}} \cite{li2025improving} & 50.3 & 50.1 & 50.0 & 50.0 & 49.9 & 50.1 & 50.0 & 50.0 & 50.1 & 50.0 & 49.9 & 50.0 & 54.7 & 98.2 & 98.5 & \underline{97.3} & 59.3 \\
C2P-CLIP \textsubscript{\textcolor{blue}{(AAAI'25)}} \cite{tan2025c2p} & 83.0 & 51.7 & 51.7 & 52.9 & 51.9 & 64.6 & 51.7 & 50.6 & 52.0 & 66.1 & 56.9 & 54.7 & 77.8 & 67.2 & 57.1 & 56.7 & 59.2  \\
AIDE \textsubscript{\textcolor{blue}{(ICLR'25)}} \cite{yan2025sanity} & 64.4 & 74.9 & 75.1 & 58.5 & 53.5 & 66.3 & 52.8 & 52.8 & 70.0 & 54.3 & 65.9 & 53.6 & 53.9 & 95.3 & 73.3 & 69.0 & 64.6  \\
DRCT \textsubscript{\textcolor{blue}{(ICML'24)}} \cite{chen2024drct} & 96.7 & 96.3 & 96.3 & 94.9 & 96.2 & 93.5 & 93.4 & 92.9 & 91.2 & 95.0 & 95.6 & 92.7 & 92.0 & 94.1 & 69.6 & 57.4 & 90.5  \\
AlignedForensics \textsubscript{\textcolor{blue}{(ICLR'25)}} \cite{rajan2025aligned} & \textbf{99.9} & \textbf{99.9} & \textbf{99.9} & \textbf{99.6} & 90.2 & 81.3 & \textbf{99.7} & 89.4 & \textbf{99.7} & 90.0 & \textbf{99.9} & \underline{99.2} & 87.6 & \textbf{99.9} & \textbf{99.8} & 92.6 & 95.5  \\
DDA \textsubscript{\textcolor{blue}{(NIPS'25)}}\cite{chen2025dual} & 99.2 & 98.9 & 99.0 & 98.3 & \underline{98.0} & \underline{96.8} & 97.9 & \underline{94.8} & 95.9 & \underline{98.2} & 98.7 & 99.0 & \underline{99.4} & 99.0 & 99.5 & 96.3 & \underline{98.1}  \\
\midrule
\textbf{LoRC (ours)} & \underline{99.6} & \underline{99.4} & \underline{99.4} & \underline{99.0} & \textbf{98.7} & \textbf{98.6} & \underline{99.4} & \textbf{99.3} & \underline{99.4} & \textbf{99.2} & \underline{99.5} & \textbf{99.5} & \textbf{99.6} & \underline{99.6} & \underline{99.6} & \textbf{99.5} & \textbf{99.3}  \\
\bottomrule
  \end{tabular}
  \end{adjustbox}
\end{table}

\begin{table}[tbp!]
\centering
\caption{Comparison of balanced accuracy on AIGCDetectionBenchmark.}
\label{tab:aigc-detection}
\begin{adjustbox}{width=1.0\linewidth}
\begin{tabular}{lcccccccccccccccccl}
\toprule
Method & ADM & DALLE2 & GLIDE & Midjourney & VQDM & BigGAN & CycleGAN & GauGAN & ProGAN & SDXL & SD14 & SD15 & StarGAN & StyleGAN & StyleGAN2 & WFR & Wukong & Avg. \\
\midrule
NPR \textsubscript{\textcolor{blue}{(CVPR'24)}} \cite{tan2024rethinking}
& 43.8 & 20.0 & 41.2 & 53.4 & 48.4 & 53.1 & 76.6 & 42.2 & 58.7 & 59.6 & 55.1 & 55.0 & 67.4 & 57.9 & 54.6 & 58.8 & 57.4 & 53.1 \\
UnivFD \textsubscript{\textcolor{blue}{(CVPR'23)}} \cite{ojha2023towards}
& 62.5 & 50.0 & 61.3 & 55.1 & 76.9 & 87.5 & \underline{96.9} & \underline{98.8} & \textbf{99.4} & 58.2 & 55.6 & 55.7 & 95.1 & 80.0 & 69.4 & 69.2 & 61.1 & 72.5 \\
FatFormer \textsubscript{\textcolor{blue}{(CVPR'24)}} \cite{liu2024forgery}
& 80.2 & 68.5 & \underline{91.1} & 54.4 & \underline{88.0} & \textbf{99.2} & \textbf{99.5} & \textbf{99.1} & 98.5 & 71.7 & 67.5 & 67.2 & \underline{99.4} & \textbf{98.0} & \textbf{98.8} & 88.3 & 75.6 & 85.0 \\
SAFE \textsubscript{\textcolor{blue}{(KDD'25)}} \cite{li2025improving}
& 49.5 & 49.5 & 53.0 & 49.0 & 50.2 & 52.2 & 51.9 & 50.0 & 50.0 & 49.8 & 49.7 & 49.8 & 50.1 & 50.0 & 50.0 & 49.8 & 50.3 & 50.3 \\
C2P-CLIP \textsubscript{\textcolor{blue}{(AAAI'25)}} \cite{tan2025c2p}
& 71.6 & 52.3 & 73.5 & 56.6 & 73.7 & \underline{98.4} & 96.8 & \underline{98.8} & \underline{99.3} & 62.3 & 77.5 & 76.9 & \textbf{99.6} & \underline{93.1} & 79.4 & \underline{94.8} & 79.4 & 81.4 \\
AIDE \textsubscript{\textcolor{blue}{(ICLR'25)}} \cite{yan2025sanity}
& 52.9 & 51.1 & 60.2 & 49.8 & 69.3 & 70.1 & 93.6 & 60.6 & 89.0 & 49.6 & 51.6 & 51.0 & 72.1 & 66.5 & 59.0 & 80.6 & 54.5 & 63.6 \\
DRCT \textsubscript{\textcolor{blue}{(ICML'24)}} \cite{chen2024drct}
& 79.9 & 89.2 & 89.2 & 85.5 & \underline{88.6} & 81.4 & 91.0 & 93.8 & 71.1 & 88.3 & 91.4 & 91.0 & 53.0 & 62.7 & 63.8 & 73.9 & 90.8 & 81.4 \\
AlignedForensics \textsubscript{\textcolor{blue}{(ICLR'25)}} \cite{rajan2025aligned}
& 51.6 & 52.0 & 55.6 & \underline{96.2} & 72.1 & 51.2 & 49.5 & 50.8 & 50.7 & \underline{95.1} & \textbf{99.7} & \textbf{99.6} & 53.8 & 52.7 & 51.6 & 50.0 & \textbf{99.6} & 66.6 \\
DDA \textsubscript{\textcolor{blue}{(NIPS'25)}}\cite{chen2025dual}
& \underline{89.5} & \underline{94.6} & 89.6 & 95.6 & 76.6 & 91.0 & 72.5 & 92.7 & 92.8 & \textbf{99.4} & 98.7 & 98.6 & 72.7 & 87.8 & 90.2 & 52.1 & \underline{98.8} & \underline{87.8} \\
\midrule
\textbf{LoRC (ours)}
& \textbf{98.4} & \textbf{99.4} & \textbf{98.5} & \textbf{97.8} & \textbf{98.8} & 93.6 & 94.9 & 95.1 & 90.6 & \textbf{99.4} & \underline{98.8} & \underline{98.8} & 89.7 & \underline{95.1} & \underline{94.3} & \textbf{95.0} & 98.7 & \textbf{96.3} \\
\bottomrule
\end{tabular}
\end{adjustbox}
\end{table}

\begin{table}[tbp!]
\centering
\caption{Comparison of balanced accuracy on Chameleon and WildRF.}
\label{tab:compare-chameleon-wildrf}
\setlength{\tabcolsep}{10pt} 
\begin{adjustbox}{width=1\linewidth}
      \begin{tabular}{l|c|cccc}
      \toprule
      \multirow{2}{*}{Method} & \multirow{2}{*}{Chameleon} & \multicolumn{4}{c}{WildRF} \\
      \cmidrule{3-6}
      &  & Facebook & Reddit & Twitter & Avg. \\
      \midrule
NPR \textsubscript{\textcolor{blue}{(CVPR'24)}} \cite{tan2024rethinking} & 59.9 & 78.1 & 61.0 & 51.3 & 63.5 \\
UnivFD \textsubscript{\textcolor{blue}{(CVPR'23)}} \cite{ojha2023towards} & 50.7 & 49.1 & 60.2 & 56.5 & 55.3 \\
FatFormer \textsubscript{\textcolor{blue}{(CVPR'24)}} \cite{liu2024forgery} & 51.2 & 54.1 & 68.1 & 54.4 & 58.9 \\
SAFE \textsubscript{\textcolor{blue}{(KDD'25)}} \cite{li2025improving} & 59.2 & 50.9 & 74.1 & 37.5 & 57.2 \\
C2P-CLIP \textsubscript{\textcolor{blue}{(AAAI'25)}} \cite{tan2025c2p} & 51.1 & 54.4 & 68.4 & 55.9 & 59.6 \\
AIDE \textsubscript{\textcolor{blue}{(ICLR'25)}} \cite{yan2025sanity} & 63.1 & 57.8 & 71.5 & 45.8 & 58.4 \\
DRCT \textsubscript{\textcolor{blue}{(ICML'24)}} \cite{chen2024drct} & 56.6 & 46.6 & 53.1 & 55.2 & 50.6 \\
AlignedForensics \textsubscript{\textcolor{blue}{(ICLR'25)}} \cite{rajan2025aligned} & 71.0 & 89.4 & 69.1 & 81.8 & 80.1 \\
DDA \textsubscript{\textcolor{blue}{(NIPS'25)}} \cite{chen2025dual} & \underline{82.4} & \underline{93.1} & \underline{86.4} & \underline{91.5} & \underline{90.3} \\
\midrule
\textbf{LoRC (ours)} & \textbf{92.6} & \textbf{98.4} & \textbf{97.7} & \textbf{96.8} & \textbf{97.6} \\
\bottomrule
      \end{tabular}
    \end{adjustbox}
\end{table}


\section{More Ablation Studies}
\label{sec:ablation}

\begin{table}[tbp!]
\centering
\small
\setlength{\tabcolsep}{5pt}
\renewcommand{\arraystretch}{1}
\caption{Ablation study on semantic decomposition. 
We evaluate the impact of low-rank attention and subspace separation with and without semantic decomposition. 
Modules are added cumulatively. 
Applying low-rank attention to entangled features (w/o decomposition) degrades performance.
Compared to simple subtraction, our orthogonal projection ensures strict geometric decoupling, preventing semantic leakage and achieving the optimal average accuracy.
}
\label{tab:ablation_decomposition}
\resizebox{\linewidth}{!}{
\begin{tabular}{lcccc}
\toprule
\textbf{Strategy} & \textbf{Standard} & \textbf{In-the-wild} & \textbf{T2I-CoReBench} & \textbf{Avg.} \\
\midrule
\multicolumn{5}{l}{\textit{w/o Semantic Decomposition}} \\
\quad Baseline (\textbf{original patch}) & 94.5 & 90.0 & 95.4 & 93.3 \\
\quad + Low-Rank Attention & 91.6 & 83.7 & 94.9 & 90.0 \\
\quad + Subspace Separation & 91.8 & 82.5 & 95.9 & 90.1 \\
\midrule
\multicolumn{5}{l}{\textit{w/ Semantic Decomposition}} \\
\quad Baseline (\textbf{subtraction}) & 94.5 & 88.7 & 97.9 & 93.7 \\
\quad + Low-Rank Attention & 94.7 & 92.1 & 95.8 & 94.2 \\
\quad + Subspace Separation & 97.0 & 94.5 & 96.6 & 96.0 \\
\midrule
\multicolumn{5}{l}{\textit{w/ Semantic Decomposition}} \\
\quad Baseline (\textbf{orthogonal}) & 97.2 & 86.9 & \textbf{98.1} & 94.1 \\
\quad + Low-Rank Attention & 96.9 & 90.2 & 96.0 & 94.4 \\
\quad + Subspace Separation & \textbf{98.1} & \textbf{95.2} & 97.0 & \textbf{96.8} \\
\bottomrule
\end{tabular}
}
\end{table}

\subsection{Effects of Semantic Decomposition}

\cref{tab:ablation_decomposition} demonstrates the critical role of semantic decomposition. Applying Low-Rank Attention directly to the original entangled features degrades the average accuracy from 93.3\% to 90.0\%. 
Without decoupling, patch tokens are dominated by high-level semantic content; consequently, the low-rank bottleneck fails to accurately capture the subtle forensic cues induced by the low-rank collapse effect.
To this end, we investigate two decomposition strategies: simple subtraction and orthogonal projection. Although simple subtraction partially mitigates feature entanglement and lifts the average accuracy to 96.0\%, it inherently suffers from semantic leakage, leaving semantic remnants in the residual space. Conversely, orthogonal projection mathematically guarantees that the isolated residuals are entirely purged of semantic directions, yielding the optimal performance of 96.8\%. 

To further verify that the decoupled residuals contain the core forensic evidence, \cref{tab:ablation-cls} ablates the contribution of global semantic information during classification.
Using only the trainable semantic \texttt{[CLS]} token yields a lower average accuracy of 93.8\%, suggesting that global semantics alone are insufficient for reliable detection.
In contrast, relying exclusively on the residual features already achieves an average accuracy of 96.2\%, and further incorporating the \texttt{[CLS]} token brings only a marginal 0.6\% gain, increasing the performance to 96.8\%.
These results confirm that the residual subspace is the primary source of detection capability, while global semantics serve only as auxiliary context.

\begin{table}[tbp!]
\centering
\small
\setlength{\tabcolsep}{5pt}
\renewcommand{\arraystretch}{1}
\caption{Ablation study on the contribution of semantic \texttt{[CLS]} token concatenation. Using only the trainable semantic \texttt{[CLS]} token yields a lower average accuracy of 93.8\%, indicating that global semantics alone are insufficient for robust detection. By comparison, residuals alone already achieve 96.2\% accuracy, and adding the frozen \texttt{[CLS]} token brings only a marginal 0.6\% gain. This confirms that the residual subspace drives detection, while global semantics merely provide auxiliary context.}
\resizebox{0.9\linewidth}{!}{
\begin{tabular}{lcccc}
\toprule
\textbf{CLS Token Strategy} & \textbf{Standard} & \textbf{In-the-wild} & \textbf{T2I-CoReBench} & \textbf{Avg.} \\
\midrule
Only Trainable \texttt{[CLS]} Token & 97.9 & 89.4 & 94.0 & 93.8 \\
w/o \texttt{[CLS]} Token & 97.1 & 94.2 & \textbf{97.3} & 96.2 \\
w/ \texttt{[CLS]} Token (LoRC) & \textbf{98.1} & \textbf{95.2} & 97.0 & \textbf{96.8} \\
\bottomrule
\end{tabular}
}
\label{tab:ablation-cls}
\end{table}

\subsection{SSL for Residual Subspace Separation}

\begin{table}[tbp!]
\centering
\small
\setlength{\tabcolsep}{5pt}
\renewcommand{\arraystretch}{1}
\caption{Ablation study on SSL placement. Applying SSL after the low-rank bottleneck achieves the best average accuracy.}
\resizebox{0.9\linewidth}{!}{%
\begin{tabular}{lcccc}
\toprule
\textbf{Setting} & \textbf{Standard} & \textbf{In-the-wild} & \textbf{T2I-CoReBench} & \textbf{Avg.} \\
\midrule
w/o SSL & 96.9 & 90.2 & 96.0 & 94.4 \\
SSL Before Low-Rank & 95.6 & 93.2 & 96.3 & 95.0 \\
SSL After Low-Rank (LoRC) & \textbf{98.1} & \textbf{95.2} & \textbf{97.0} & \textbf{96.8} \\
\bottomrule
\end{tabular}%
}
\label{tab:ablation-ssl-placement}
\end{table}
We conduct an ablation study on the placement of SSL to examine its interaction with the low-rank bottleneck. As shown in Tab.~\ref{tab:ablation-ssl-placement}, Low-Rank Attention provides the rank bottleneck, while SSL further improves the separation of real/fake residual distributions.
Placing SSL after the low-rank bottleneck allows nuisance factors such as resizing and compression artifacts to be suppressed before subspace separation. As a result, SSL operates on cleaner residuals and better separates real/fake covariance supports. In contrast, applying SSL before the low-rank bottleneck may over-separate entangled nuisance factors in the unfiltered residual space, especially on the Standard Benchmark where early-generator artifacts are often coupled with content biases. This explains why SSL after Low-Rank Attention achieves the best Standard accuracy of 98.1\% and the best average accuracy of 96.8\%.

\subsection{Impact of the Vision Backbone}

\cref{tab:ablation-backbone3} ablates the vision backbone. Our framework scales consistently across the DINOv3 family, improving from 93.6\% with DINOv3-L to 97.4\% with DINOv3-7B. We attribute this trend to the stronger semantic representations of larger DINOv3 models, which allow our orthogonal projection to remove semantic components and isolate forensic artifacts more cleanly. Although DINOv3-7B achieves the best accuracy at 97.4\%, we adopt DINOv3-H+ (96.8\%) as the default extractor to better balance performance and computational overhead.

\begin{table}[tpb!]
\centering
\small
\setlength{\tabcolsep}{5pt}
\renewcommand{\arraystretch}{1}
\caption{Ablation study of LoRC across different DINOv3 backbones.}
\resizebox{0.9\linewidth}{!}{
\begin{tabular}{llcccccc}
\toprule
\textbf{Backbone} & \textbf{Variant} & \textbf{Standard} & \textbf{In-the-wild} & \textbf{T2I-CoReBench} & \textbf{Avg.} & \textbf{Latency (ms)} \\
\midrule
DINOv3-L  & Original Patch & 92.8   & 82.7   & 94.3   & 89.9   & \multirow{2}{*}{5.4} \\
          & LoRC           & 96.5   & 88.1   & 96.3   & 93.6   &  \\
\midrule
DINOv3-H+ & Original Patch & 94.5   & 90.0   & 95.4   & 93.3   & \multirow{2}{*}{9.5} \\
          & LoRC           & \textbf{98.1} & 95.2 & 97.0 & 96.8 &  \\
\midrule
DINOv3-7B & Original Patch & 96.1   & 92.9   & 96.5   & 95.2   & \multirow{2}{*}{85.4} \\
          & LoRC           & 97.6 & \textbf{96.4} & \textbf{98.3} & \textbf{97.4} &  \\
\bottomrule
\end{tabular}
}
\label{tab:ablation-backbone3}
\end{table}

\section{Inference Efficiency Analysis}
\label{sec:computational-cost}

\begin{figure}[tb!]
  \centering
  \includegraphics[width=\linewidth]{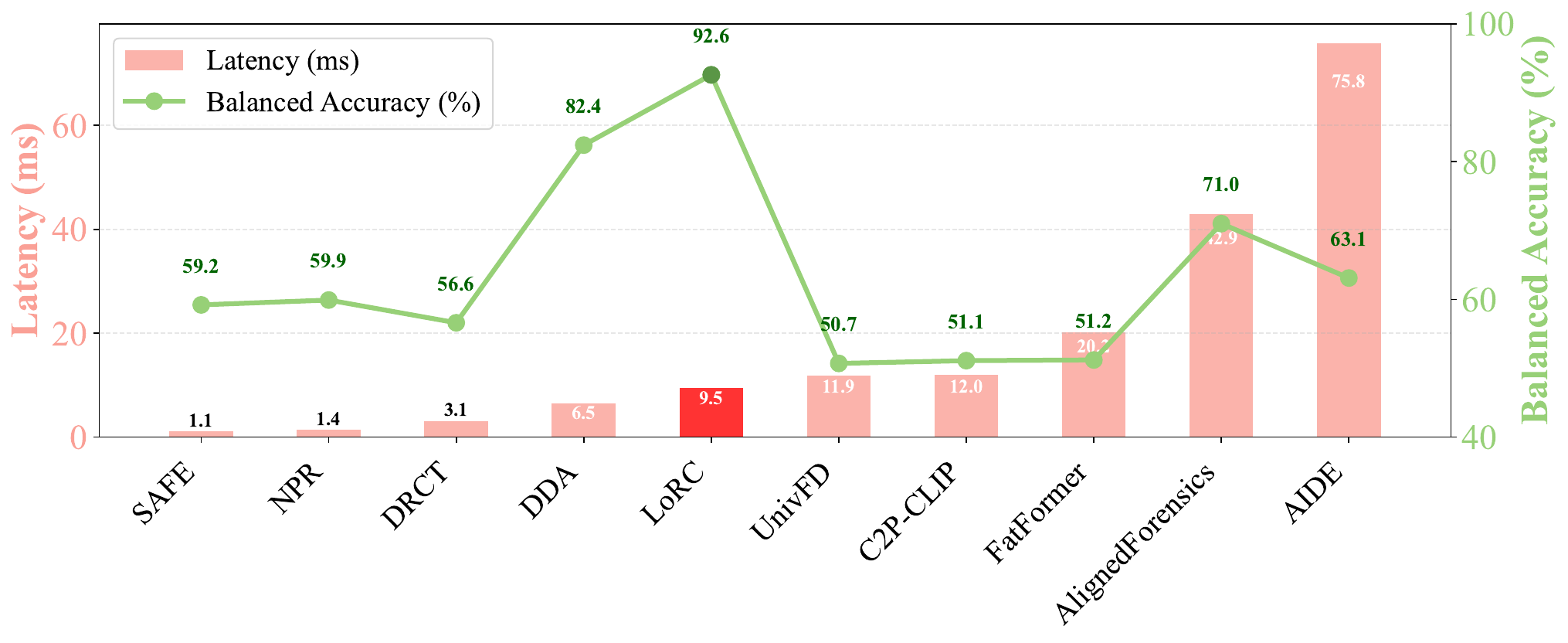}
  \caption{
    Efficiency and accuracy comparison on the Chameleon dataset. LoRC achieves a strong balance between accuracy and efficiency, attaining 92.6\% balanced accuracy with a throughput of 106 images/s (9.5\,ms latency).
  }
  \label{fig:inference_time}
\end{figure}

We further evaluate the inference efficiency of different methods. Specifically, all methods are tested on the full Chameleon dataset with a batch size of 16, under their respective inference settings. The results are shown in \cref{fig:inference_time}. 
LoRC maintains 92.6\% balanced accuracy while achieving 9.5\,ms latency and 106 images/s throughput, demonstrating a favorable balance between accuracy and efficiency.

\end{document}